\documentclass{article}

\usepackage{arxiv}
\usepackage[utf8]{inputenc}
\usepackage[T1]{fontenc}
\usepackage{url}
\usepackage{booktabs}
\usepackage{amsfonts}
\usepackage{nicefrac}
\usepackage{microtype}
\usepackage{graphicx}
\usepackage{natbib}
\usepackage{doi}
\usepackage{xcolor}
\usepackage{subcaption}
\usepackage{amsmath}
\usepackage{multirow}
\usepackage{authblk}
\usepackage{hyperref}

\newcommand{\suppfig}[1]{\hyperref[#1]{Supplementary Figure~\ref*{#1}}}
\newcommand{\supptab}[1]{\hyperref[#1]{Supplementary Table~\ref*{#1}}}

\newcommand{\beginsupplementary}{%
  \clearpage
  \setcounter{figure}{0}
  \setcounter{table}{0}
  \captionsetup[figure]{name=Supplementary Figure}
  \captionsetup[table]{name=Supplementary Table}
  \renewcommand{\thefigure}{\arabic{figure}}
  \renewcommand{\thetable}{\arabic{table}}
  \renewcommand{\figureautorefname}{Supplementary Figure}
  \renewcommand{\tableautorefname}{Supplementary Table}
}

\title{\emph{GraphDerm}: Fusing Imaging, Physical Scale, and Metadata in a Population-Graph Classifier for Dermoscopic Lesions}

\author[1,2]{Mehdi Yousefzadeh\textsuperscript{*}\thinspace}
\author[1]{Parsa Esfahanian\textsuperscript{*}\thinspace}
\author[1]{Sara Rashidifar}
\author[3]{Hossein Salahshoor Gavalan}
\author[4]{\\Negar Sadat Rafiee Tabatabaee}
\author[5]{Saeid Gorgin}
\author[6]{Dara Rahmati}
\author[7]{Maryam Daneshpazhooh\textsuperscript{†}\thinspace}

\affil[1]{School of Computer Science, Institute for Research in Fundamental Sciences (IPM), Tehran, Iran}
\affil[2]{Department of Physics, Shahid Beheshti University, Tehran, Iran}
\affil[3]{Faculty of Science, University of Tehran, Tehran, Iran}
\affil[4]{Fatemeh Zahra Hospital, Alborz University of Medical Sciences, Eshtehard, Iran}
\affil[5]{Department of Electrical and Computer Engineering, Sungkyunkwan University, Suwon, South Korea}
\affil[6]{Department of Computer Science and Engineering, Shahid Beheshti University, Tehran, Iran}
\affil[7]{Autoimmune Bullous Diseases Research Center, Razi Hospital, Tehran University of Medical Sciences, Tehran, Iran}

\affil[ ]{\small \texttt{
\href{mailto:yousefzadeh.meh@gmail.com}{yousefzadeh.meh@gmail.com},
\href{mailto:parsa.esfahanian@ipm.ir}{parsa.esfahanian@ipm.ir},
\href{mailto:rashidifar.sara@gmail.com}{rashidifar.sara@gmail.com},
\href{mailto:hoseinsalahshoor1990@gmailOk.com}{hoseinsalahshoor1990@gmail.com},
\href{mailto:nsrafiee425@gmail.com}{nsrafiee425@gmail.com},
\href{mailto:gorgin81@skku.edu}{gorgin81@skku.edu},
}}
\affil[ ]{\small \texttt{
\href{mailto:d_rahmati@sbu.ac.ir}{d\_rahmati@sbu.ac.ir},
\href{mailto:maryamdanesh.pj@gmail.com}{maryamdanesh.pj@gmail.com}
}}
\renewcommand{\shorttitle}{\emph{GraphDerm}: Fusing Imaging, Physical Scale, and ...}

\hypersetup{
    pdftitle={GraphDerm: Fusing Imaging, Physical Scale, and Metadata in a Population-Graph Classifier for Dermoscopic Lesions},
    pdfsubject={cs.LG, q-bio.QM},
    pdfauthor={Mehdi Yousefzadeh, Parsa Esfahanian, Sara Rashidifar, Hossein Salahshoor Gavalan, Negar Sadat Rafiee Tabatabaee, Saeid Gorgin, Dara Rahmati, Maryam Daneshpajouh},
    pdfkeywords={Graph Neural Network, Dermoscopy, Lesion Segmentation, Ruler Segmentation, Scale Calibration},
}

\begin{document}
\maketitle

\renewcommand{\thefootnote}{\fnsymbol{footnote}}
\setcounter{footnote}{1}
\footnotetext[1]{Equal contribution.}
\footnotetext[2]{Corresponding author.}

\begin{abstract}
\textbf{Introduction.}
Dermoscopy aids melanoma triage, yet image-only AI often ignores patient metadata (age, sex, site) and the physical scale needed for geometric analysis. We present \emph{GraphDerm}, a population-graph framework that fuses imaging, millimeter-scale calibration, and metadata for multiclass dermoscopic classification, to the best of our knowledge the first ISIC-scale application of GNNs to dermoscopy.

\textbf{Methods.}
We curate ISIC 2018/2019, synthesize ruler-embedded images with exact masks, and train U-Nets (SE-ResNet-18) for lesion and ruler segmentation. Pixels-per-millimeter are regressed from the ruler-mask two-point correlation via a lightweight 1D-CNN. From lesion masks we compute real-scale descriptors (area, perimeter, radius of gyration). Node features use EfficientNet-B3; edges encode metadata/geometry similarity (fully weighted or thresholded). A spectral GNN performs semi-supervised node classification; an image-only ANN is the baseline.

\textbf{Results.}
Ruler and lesion segmentation reach Dice $0.904$ and $0.908$; scale regression attains MAE $1.5$ px (RMSE $6.6$). The graph attains AUC $0.9812$, with a thresholded variant using $\sim25\%$ of edges preserving AUC $0.9788$ (vs.\ $0.9440$ for the image-only baseline); per-class AUCs typically fall in the $0.97$–$0.99$ range.

\textbf{Conclusion.}
Unifying calibrated scale, lesion geometry, and metadata in a population graph yields substantial gains over image-only pipelines on ISIC-2019. Sparser graphs retain near-optimal accuracy, suggesting efficient deployment. Scale-aware, graph-based AI is a promising direction for dermoscopic decision support; future work will refine learned edge semantics and evaluate on broader curated benchmarks.
\end{abstract}


\label{sec:introduction}
\section{Introduction}

Skin cancer is the most common malignancy worldwide and represents a major public-health challenge. The World Health Organization (WHO) estimates between two to three million new non-melanoma skin cancers (NMSC) and more than 300{,}000 new cases of melanoma each year, with melanoma alone accounting for over 57{,}000 deaths annually \cite{sung2021global, arnold2022global}. Although basal cell carcinoma (BCC) and squamous cell carcinoma (SCC) comprise the majority of skin malignancies, melanoma is far more aggressive and disproportionately responsible for mortality \cite{whiteman2016growing}. Its incidence continues to rise, particularly in fair-skinned populations and regions with high ultraviolet exposure \cite{web1}, underscoring the urgent need for effective screening and timely intervention. Late-stage diagnosis is strongly associated with poor five-year survival, highlighting the clinical imperative of early detection and precise triage \cite{jones2019dermoscopy}.

Despite its importance, reliable melanoma detection remains challenging. Clinical examination alone is limited by the resemblance of malignant to benign pigmented lesions, as well as similarities to non-cancerous dermatological disorders. To standardize assessment, dermatologists commonly employ the \emph{ABCDE} rule (Asymmetry, Border irregularity, Color variegation, Diameter, Evolving lesion) as a mnemonic for suspicious features \cite{nachbar1994abcd, friedman1985clinical}. Lesions larger than 6\,mm in diameter are particularly concerning for melanoma. Yet in practice, diagnostic accuracy varies substantially: biopsy accuracy ranges between 49\% and 81\%, and up to one-third of melanomas are initially misclassified as benign\cite{mackenzie1998melanoma, grin1990accuracy}.

Dermoscopy (dermatoscopy) has therefore become the \emph{de facto} non-invasive imaging modality for evaluating skin lesions in routine practice. By revealing sub-macroscopic structures, dermoscopy improves both sensitivity and specificity compared to naked-eye inspection, with reported diagnostic accuracies of 75–97\% in experienced hands \cite{kittler2002dermoscopy, dinnes2018dermoscopy, saurat2004dermoscopy, celebi2005unsupervised}. Nevertheless, performance is highly operator-dependent: manual interpretation is time-consuming, error-prone, and subject to inter- and intra-observer variability, particularly among less experienced clinicians \cite{jones2019dermoscopy}. These limitations have accelerated the development of Computer-Assisted Diagnosis (CAD) systems to support dermatologists.

Over the last decade, deep learning has transformed medical image analysis, including skin cancer detection. Convolutional Neural Networks (CNNs) have achieved dermatologist-level performance in binary melanoma classification tasks \cite{esteva2017dermatologist, hekler2019superior}, and dominate lesion segmentation and classification challenges such as ISIC 2017–2019 \cite{codella2018skin, codella2019skin, tschandl2018ham10000, combalia2019bcn20000}. Models leveraging architectures such as U-Net \cite{ronneberger2015u}, DenseNet \cite{huang2017densely}, and their variants have advanced lesion segmentation \cite{yuan2017improving, yu2016automated}, while pipelines combining segmentation and classification have demonstrated state-of-the-art performance \cite{lee2018wonderm}. More recent efforts have explored optimization-based dual networks \cite{gomathi2023skin}, hybrid CNN–RF classifiers \cite{mustafa2025deep}, and efficient backbones such as MobileNetV3 \cite{kumar2024precise}, reporting accuracies approaching 98\% on HAM10000.  

Despite these advances, several challenges remain unresolved. First, multiclass lesion classification is substantially more difficult than binary melanoma detection, due to extreme class imbalance and subtle inter-class variability. Second, progress is constrained by limited dataset size, heterogeneous acquisition protocols, and the high cost of expert annotations. Transfer learning and domain adaptation strategies \cite{wang2019domain} partially mitigate these issues but have limited generalization in real-world clinical deployment. Finally, many CNN pipelines rely solely on pixel intensities, ignoring clinically salient cues such as lesion geometry and patient context. These gaps motivate exploration of richer modeling paradigms. 

Most existing CNN-based methods treat dermoscopic images in isolation, neglecting two critical elements of clinical reasoning: (i) auxiliary metadata such as age, sex, and anatomic site, which contextualize disease likelihood, and (ii) physical scale, essential for quantifying geometric features such as the diameter criterion in the ABCDE rule \cite{nachbar1994abcd}. Importantly, physical scale cannot be inferred from raw pixels alone and is often absent due to heterogeneous imaging protocols and the lack of calibration markers \cite{codella2018skin, codella2019skin, jones2019dermoscopy, gandhi2015skin}.  

Population graphs provide a principled way to incorporate these factors. In this paradigm, each patient or lesion is represented as a node with image-derived features, while edges encode inter-case similarity based on metadata or geometric cues. Graph Neural Networks (GNNs) then enable semi-supervised learning and label propagation across the cohort, leveraging relationships that extend beyond individual pixels \cite{kipf2016semi}.  

In the broader medical imaging literature, population-graph GNNs have demonstrated substantial gains. \cite{parisot2018disease} applied GCNs to neuroimaging cohorts for autism spectrum disorder (ABIDE) and Alzheimer’s disease (ADNI), showing that integrating phenotypic metadata with MRI features improves classification over image-only models \cite{parisot2018disease}. Subsequent work extended these approaches to EEG analysis \cite{song2018eeg}, cerebral cortex parcellation \cite{gopinath2020graph}, anatomical segmentation \cite{noh2020combining}, and multi-modal disease prediction\cite{rakhimberdina2020population,huang2017densely}. Collectively, these studies highlight the promise of GNNs for capturing population structure and context—yet dermatology remains underexplored in this respect.

In this work, we propose \emph{GraphDerm}, a population-graph framework for multiclass dermoscopic lesion classification. To the best of our knowledge, this is the first application of GNNs to ISIC-scale dermoscopy. The framework fuses image-derived features with explicit physical scale information and patient metadata, addressing a key limitation of conventional CNN-only pipelines that operate on images in isolation.  

To enable scale-aware modeling, we curate the ISIC 2019 dataset by systematically identifying images with embedded rulers and synthesizing ruler-bearing dermoscopic images from ruler-free sources. This provides exact ruler masks by construction, which serve as supervision for a dedicated ruler-segmentation module. From these masks, we estimate pixels-per-millimeter using a two-point correlation function (TPCF) signature regressed by a lightweight CNN, yielding accurate lesion geometry in physical units.  

Building on this, we compute real-scale lesion descriptors including area, perimeter, and a radius of gyration in millimeters directly from segmentation masks. These geometric features are then combined with auxiliary metadata such as age, sex, anatomic site, and dataset source to define the similarity structure of the population graph. We systematically explore multiple edge construction and sparsification strategies, including fully weighted, thresholded, random, and identical edge schemes. The fully weighted graph achieves the best performance with AUC $=0.9812$, while a thresholded variant at $T{=}0.7$ retains nearly identical AUC $=0.9788$ using only about 25\% of edges.  

Across all experimental settings, neighborhoods informed by metadata and real-scale geometry consistently outperform image-only CNN baselines on ISIC 2019. These results demonstrate that integrating physical scale and cohort structure into dermoscopic classification yields measurable gains over conventional pipelines and highlights the promise of graph-based approaches for clinical dermatology.  

\paragraph{Contributions.} In short, the contributions of this research are:
\begin{itemize}
  \item \textbf{Scale-aware population graph.} We introduce \emph{GraphDerm}, a framework that unifies image features, calibrated lesion geometry, and patient metadata within a single graph-based model.
  \item \textbf{Calibration and geometry pipeline.} We develop a practical procedure to obtain ruler signals and segmentation masks (including synthetic augmentation), estimate physical scale, and derive real-scale geometric descriptors.
  \item \textbf{Graph design and sparsification.} We investigate alternative neighborhood constructions and sparsity levels, showing that carefully structured sparse graphs retain performance close to dense counterparts.
  \item \textbf{Empirical validation.} We demonstrate consistent gains over image-only baselines on a large multiclass dermoscopy benchmark, with stable behavior across classes.
\end{itemize}

This paper is organized as follows. Section~\ref{sec:method} details dataset curation, ruler synthesis, segmentation models, the TPCF-based scale estimator, real-scale geometric feature extraction, and graph construction and GNN design. Section~\ref{sec:results} reports segmentation and scale-estimation accuracy, ablations over edge construction and sparsification, and end-to-end classification performance. Section~\ref{sec:discussion} examines the implications and limitations of our approach, including learned edge semantics and broader benchmarks. And finally, section~\ref{sec:conclusion} synthesizes the key findings, final takeaways, and avenues for clinical translation.

\section{Methods} \label{sec:method}

This section details the \emph{GraphDerm} pipeline for multiclass dermoscopic lesion classification, integrating image analysis, physical calibration, and cohort-level context. The framework is designed to address the limitations of conventional CNN-only pipelines by explicitly modeling real-scale lesion geometry and patient metadata within a graph-based framework.

We first curate dermoscopic corpora from the ISIC 2018 and ISIC 2019 challenges to form the working dataset. From ruler-free images, we synthesize ruler-bearing counterparts with paired ruler masks, which enable supervised training of dedicated U-Net models for ruler and lesion segmentation. Predicted ruler masks are then used to estimate millimeters-per-pixel via a two-point correlation function regressed by a lightweight CNN, yielding precise calibration of physical scale. With this calibration, we compute lesion geometric descriptors including area, perimeter, and  radius of gyration in millimeters.

Finally, we construct a population graph where each node encodes imaging-derived features together with scale-aware geometry, and edges capture inter-patient similarity defined by metadata such as age, sex, anatomic site, and dataset source. Multiple edge-weighting schemes are investigated—fully weighted, thresholded, random, and identical—before applying a spectral Graph Neural Network (GNN) for semi-supervised multiclass classification.

\subsection{Datasets} \label{sec:dataset}
The \emph{GraphDerm} dataset was constructed by curating, merging, and processing two public dermoscopy cohorts: the 2018 ISIC challenge set \cite{tschandl2018ham10000, codella2019skin} and the 2019 ISIC challenge set \cite{tschandl2018ham10000, combalia2019bcn20000}. \emph{GraphDerm} contains synthesized dermoscopic images with embedded rulers together with pixel-wise segmentation masks for the skin lesion and the ruler, plus associated metadata.

\begin{figure}[!b]
    \centering
    \includegraphics[width=\textwidth]{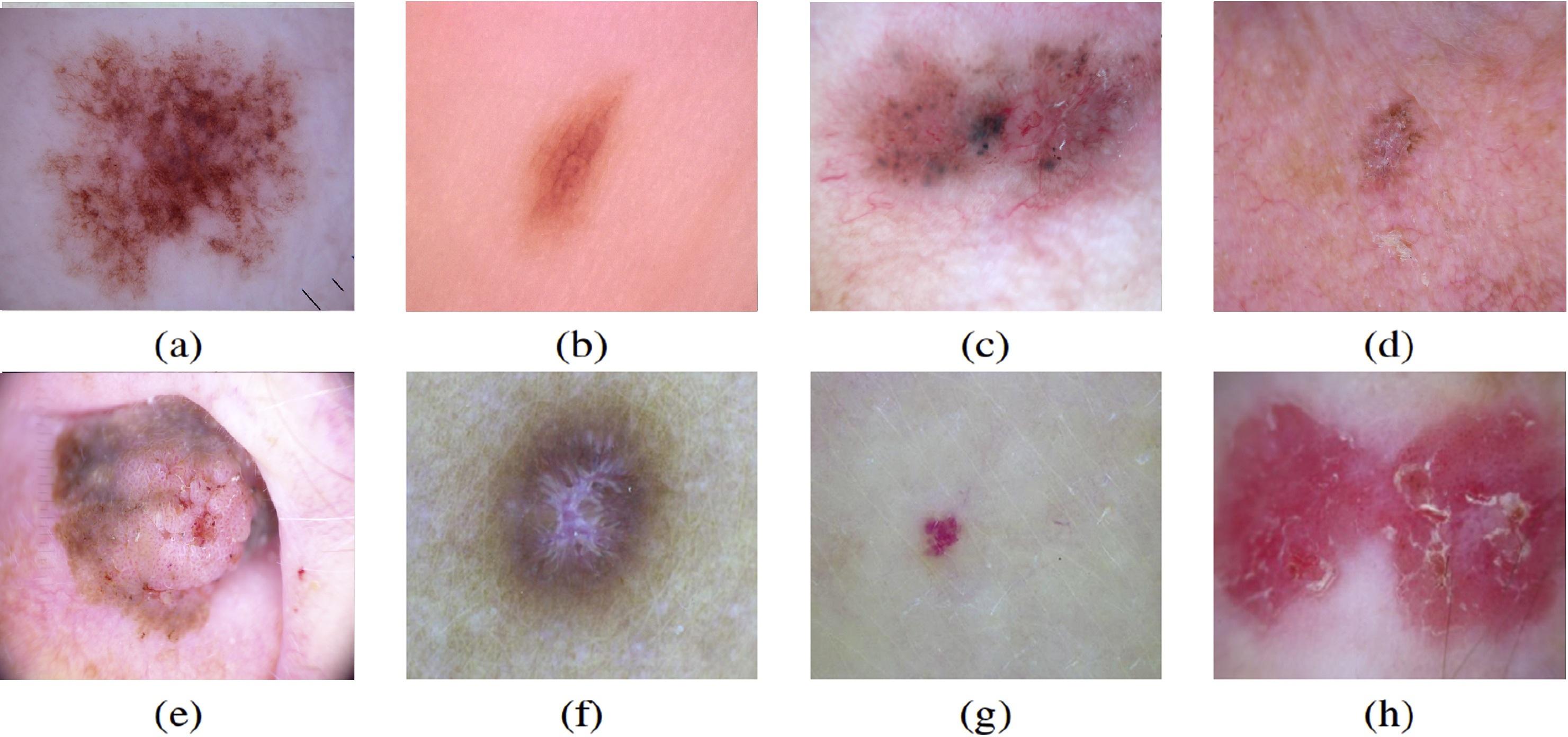}
    \caption{
        One sample image per each of the 8 diagnostic classes in the \emph{GraphDerm} dataset:
        (a) Melanoma, (b) Melanocytic Nevus, (c) Basal Cell Carcinoma, (d) Actinic Keratosis, 
        (e) Benign Keratosis, (f) Dermatofibroma, (g) Vascular Lesion, and (h) Squamous Cell Carcinoma.
    }
    \label{fig:classes}
\end{figure}

Within \emph{GraphDerm}, the ISIC 2018 and ISIC 2019 data serve two complementary goals: lesion segmentation and ruler segmentation, respectively. The dataset includes eight diagnostic classes; their distribution is reported in \supptab{table:isic2019-trainval}, and one representative example from each class is shown in \suppfig{fig:classes}.

\begin{figure}[!b]
    \centering
    \includegraphics[width=0.8\textwidth]{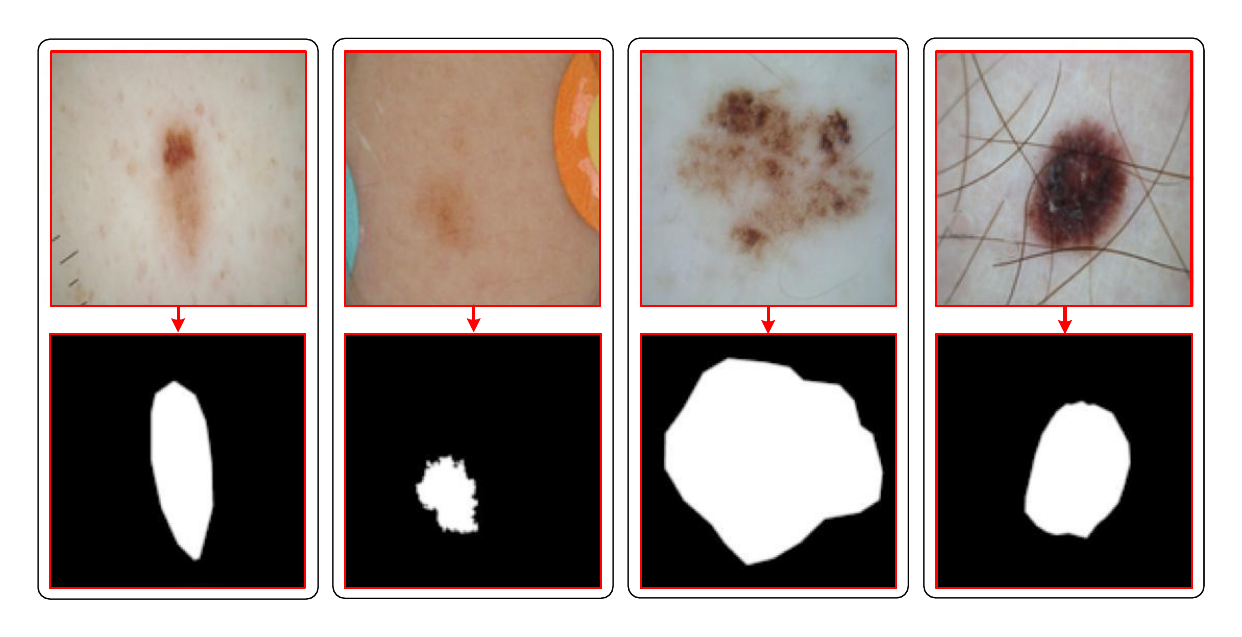}
    \caption{
        Example dermoscopic images from ISIC 2018 Task~1 alongside their provided lesion segmentation masks used as ground truth.
    }
    \label{fig:isic2018}
\end{figure}

\paragraph{Lesion segmentation data (ISIC 2018).}
The ISIC 2018 Task~1 dataset comprises 2{,}594 dermoscopic images, each accompanied by a suggested lesion segmentation mask that we use as ground truth. Examples are provided in \autoref{fig:isic2018}. A U-Net model \cite{ronneberger2015u} with an encoder pretrained on ImageNet \cite{} was trained on this set for lesion mask prediction. Standard dermoscopy preprocessing and augmentations were applied.

\paragraph{Source pool for ruler work (ISIC 2019).}
To preserve acquisition consistency, we retained the 25{,}331 raw dermoscopic images from ISIC 2019 across the eight diagnostic classes listed in \supptab{table:isic2019-trainval}. 

\begin{figure}[!t]
    \centering
    \includegraphics[width=\textwidth]{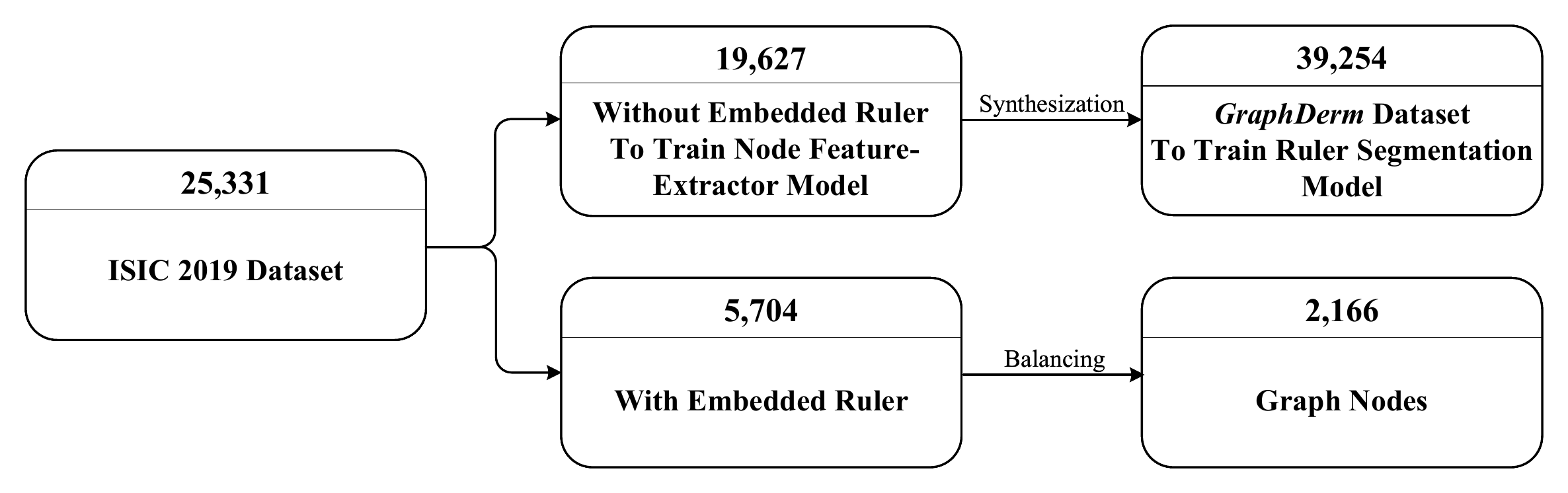}
    \caption{
        Taxonomy of how ISIC 2019 is utilized within the \emph{GraphDerm} framework.
    }
    \label{fig:taxonomy}
\end{figure}

Images that already contained visible embedded rulers but lacked ruler masks were treated as noisy and removed via manual screening. After this cleaning, 19{,}627 ruler-free images remained and were reserved for ruler synthesis (see \suppfig{fig:rulers&mms}). Among the remaining 5{,}704 images, class counts were imbalanced; we therefore kept all minority-class samples and subsampled the majority classes to approximately balance labels. A subset of 2{,}166 images was further used to construct the population graph; the overall taxonomy and usage of ISIC 2019 within \emph{GraphDerm} are summarized in \autoref{fig:taxonomy}.

\subsection{Synthetic Data Generation for Ruler Segmentation}
\begin{figure}[!ht]
    \centering
    \begin{subfigure}{0.5\textwidth}
        \centering
        \includegraphics[width=\textwidth]{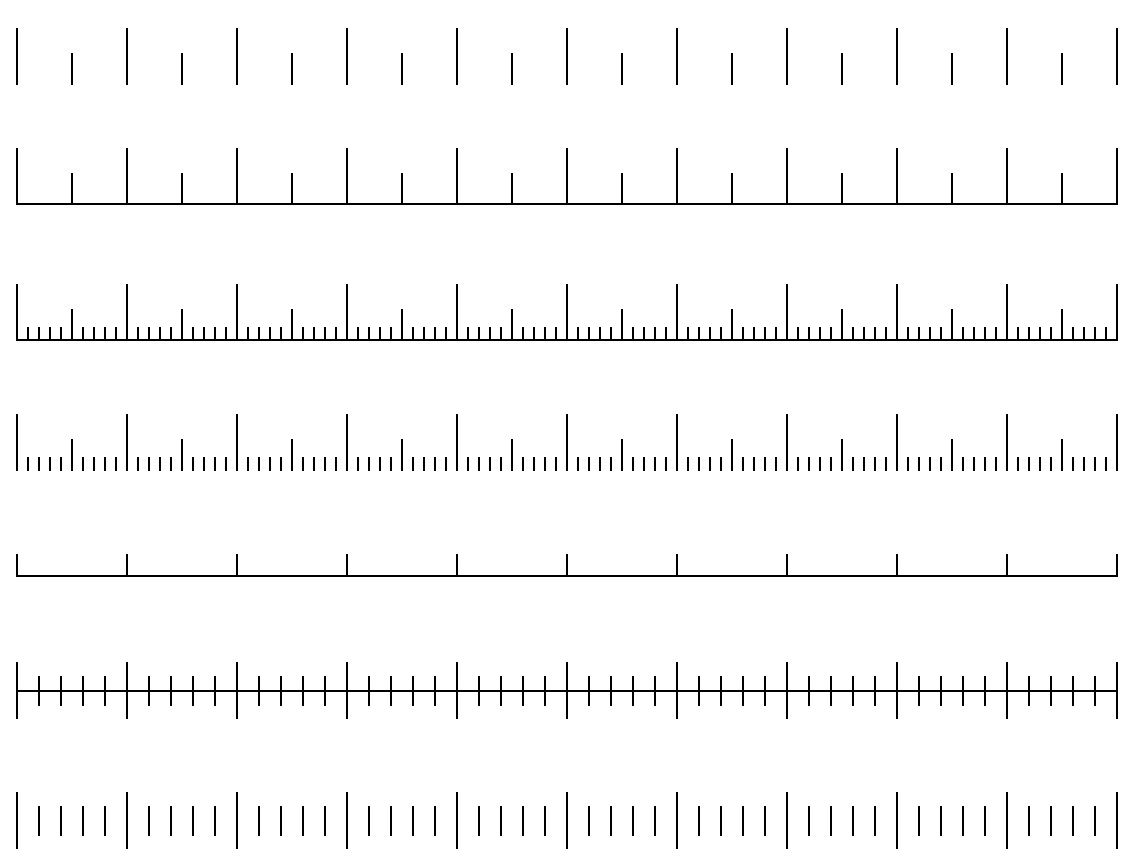}
        \label{fig:rulers}
    \end{subfigure}
    \hfill
    \begin{subfigure}{0.5\textwidth}
        \centering
        \includegraphics[width=\textwidth]{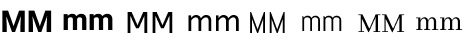}
        \label{fig:mms}
    \end{subfigure}
    \caption{
    Custom-designed rulers and millimeter markers used for embedding into synthesized dermoscopic images.
    }
    \label{fig:rulers&mms}
\end{figure}

For training a robust ruler-segmentation model, we synthesized ruler-bearing dermoscopy images from the 19{,}627 cleaned ruler-free ISIC 2019 images. We designed seven custom rulers and eight millimeter markers (\suppfig{fig:rulers&mms}); each ruler image was stored as a transparent PNG and resized to $512\times512$ prior to embedding.

\paragraph{Synthesis pipeline.}
Each synthesized image was produced by the following stochastic procedure:
\begin{enumerate}
    \small
    \item Given that real ruler appearances most closely match the top ruler in \suppfig{fig:rulers&mms}, we select that ruler with probability $50\%$; the other six rulers share the remaining $50\%$ uniformly.
    \item With probability $20\%$, one of the millimeter markers in \suppfig{fig:rulers&mms} is selected uniformly at random and overlaid on the ruler.
    \item A contiguous segment of the chosen ruler is cropped along its length to a random fraction between $30\%$ and $100\%$ of its original length.
    \item The resulting ruler segment is embedded onto the dermoscopic image at a random position.
    \item With probability $50\%$, a black circular occlusion mask is added to the image.
    \item Additive white noise with random intensity drawn from a uniform distribution is applied to the image.
    \item With probability $50\%$, the entire image is rotated by a random angle in $[0^\circ, 360^\circ]$ chosen uniformly.
    \item The image is scaled up or down uniformly at random by $20\%$.
    \item A Gaussian filter with $\sigma$ sampled from a normal distribution and clipped to $[0.5, 5.5]$ is applied.
    \item The synthesized image is accepted only if at least $20\%$ of ruler line pixels (black ticks) remain after transformations; otherwise the process is repeated.
\end{enumerate}

This pipeline was executed twice per source image, yielding 39{,}254 ruler-embedded synthetic images in total. Because the embedding transformations are controlled, we also generate the corresponding ruler segmentation masks at step~4; any global transforms in steps~7 and 8 are applied identically to the masks. The synthesized images closely mimic real cases while providing precise, high-quality ruler masks.

\paragraph{Ruler-segmentation model and training.}
Two U-Net variants were evaluated with MobileNetV2 \cite{Sandler_2018_CVPR} and SE-ResNet-18 \cite{Hu_2018_CVPR} backbones; the latter performed best and was adopted. Due to the importance of spatial detail for thin ruler ticks, inputs were $512\times512$ with batch size $16$. Models were trained for $60$ epochs using Adam \cite{kingma2014adam} with an initial learning rate of $2\mathrm{e}{-3}$ decayed over epochs. Loss function was a custom sum of binary cross-entropy and Dice:
\begin{equation}
    \mathrm{BCE} + \mathrm{Dice} = 
    \Big(-\frac{1}{N} \sum_{i=1}^{N} [T_i\log(P_i) + (1 - T_i)\log(1 - P_i)]\Big)
    + \Big(1 - \frac{2 \times P \cap T}{P + T}\Big),
    \label{eq:bce+dice}
\end{equation}
where $P$ and $T$ denote the predicted and true pixel labels, respectively.

\subsection{Statistical and Geometric Properties}
\begin{figure}[!b]
    \centering
    \includegraphics[width=\textwidth]{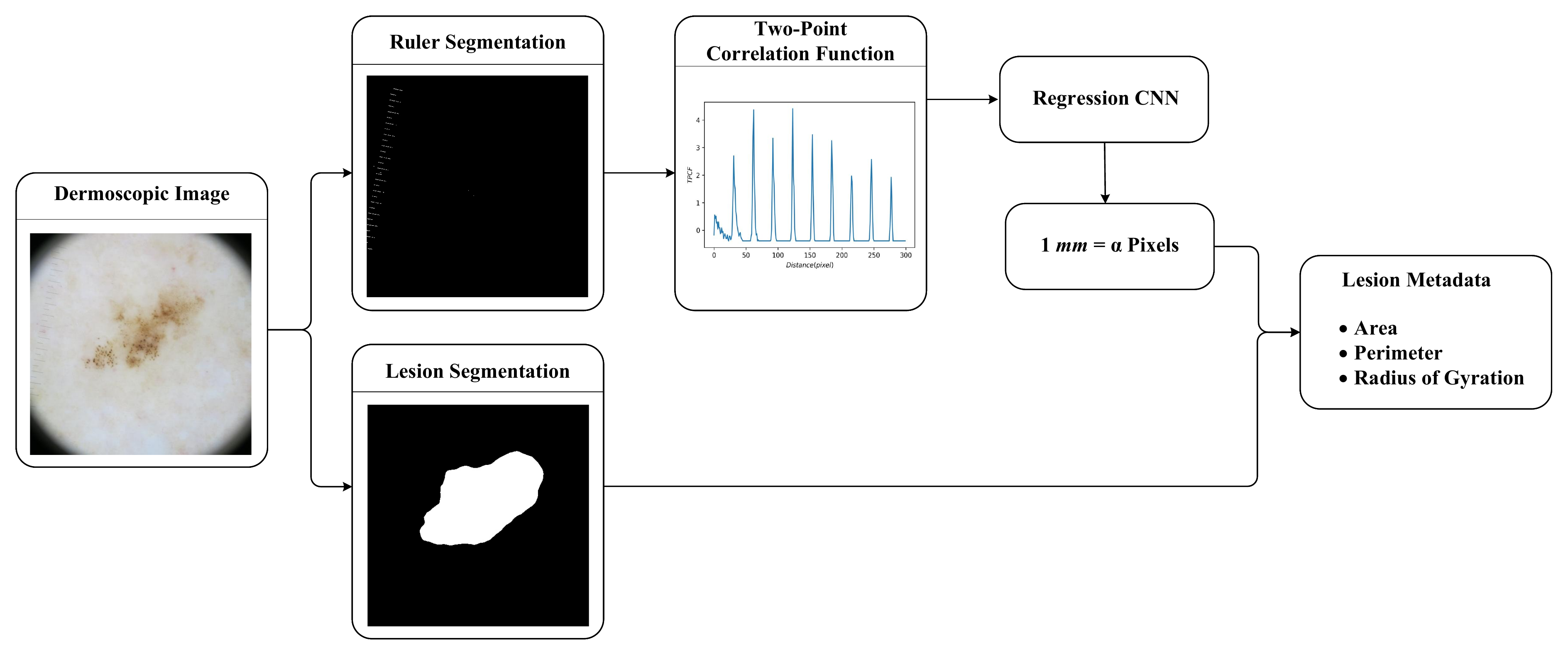}
    \caption{
        Overview of the \emph{GraphDerm} pipeline. From each dermoscopic image, ruler and lesion segmentation masks are predicted. The two-point correlation function on the ruler mask is input to a CNN regressor that estimates pixels-per-millimeter. Combining this scale with the lesion mask yields geometric lesion features.
    }
    \label{fig:GraphDerm}
\end{figure}

To support screening and quantification, lesion descriptors such as area, perimeter, and a  radius of gyration are derived in physical units (\emph{mm}). This is done by first estimating the pixel scale via a two-point correlation analysis of the predicted ruler mask, and then applying geometric computations to the predicted lesion mask. An overview of the \emph{GraphDerm} pipeline is shown in \autoref{fig:GraphDerm}.

\begin{figure}[!h]
    \centering
    \begin{subfigure}{0.49\textwidth}
        \centering
        \includegraphics[width=\textwidth]{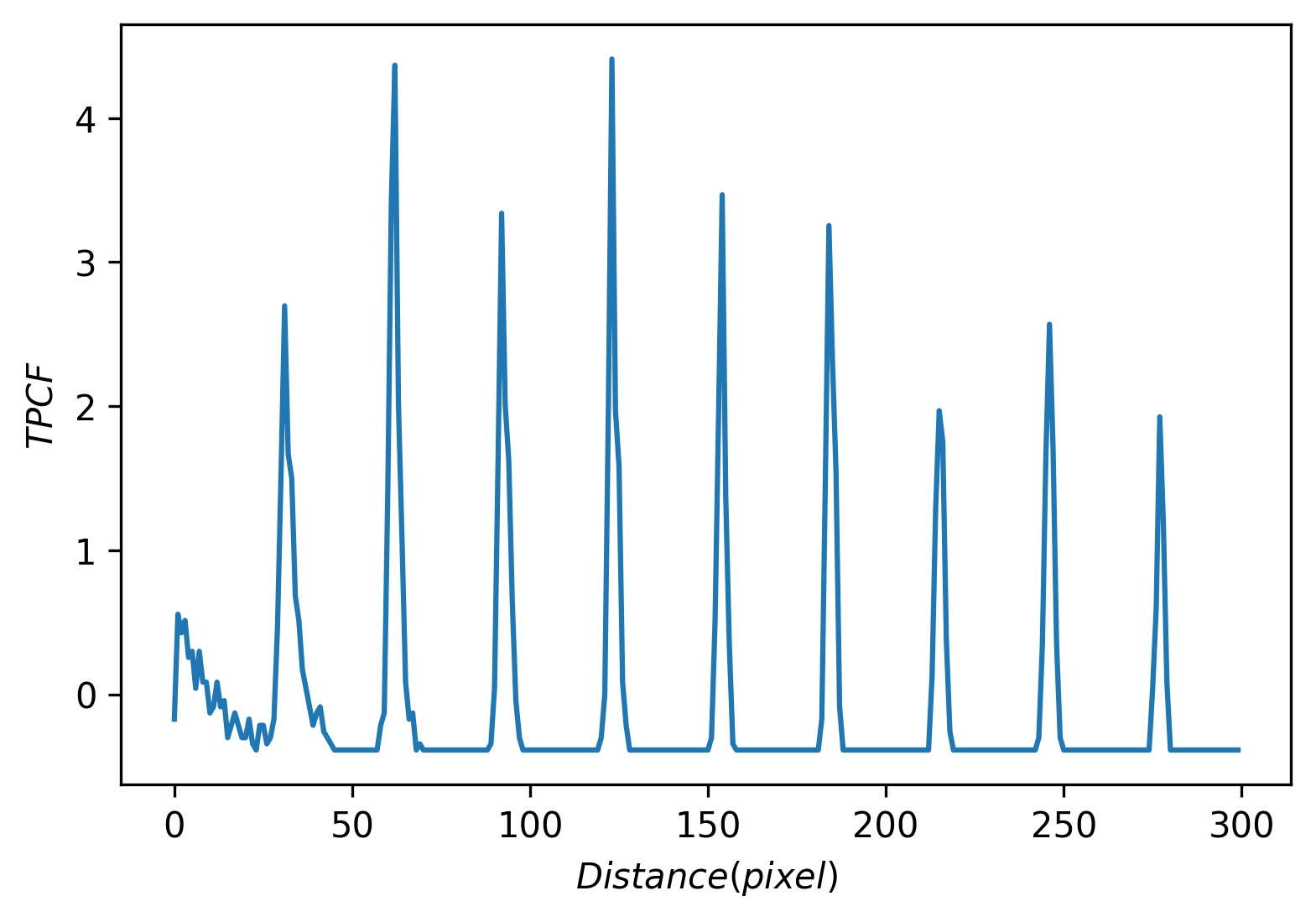}
    \end{subfigure}
    \hfill
    \begin{subfigure}{0.49\textwidth}
        \centering
        \includegraphics[width=\textwidth]{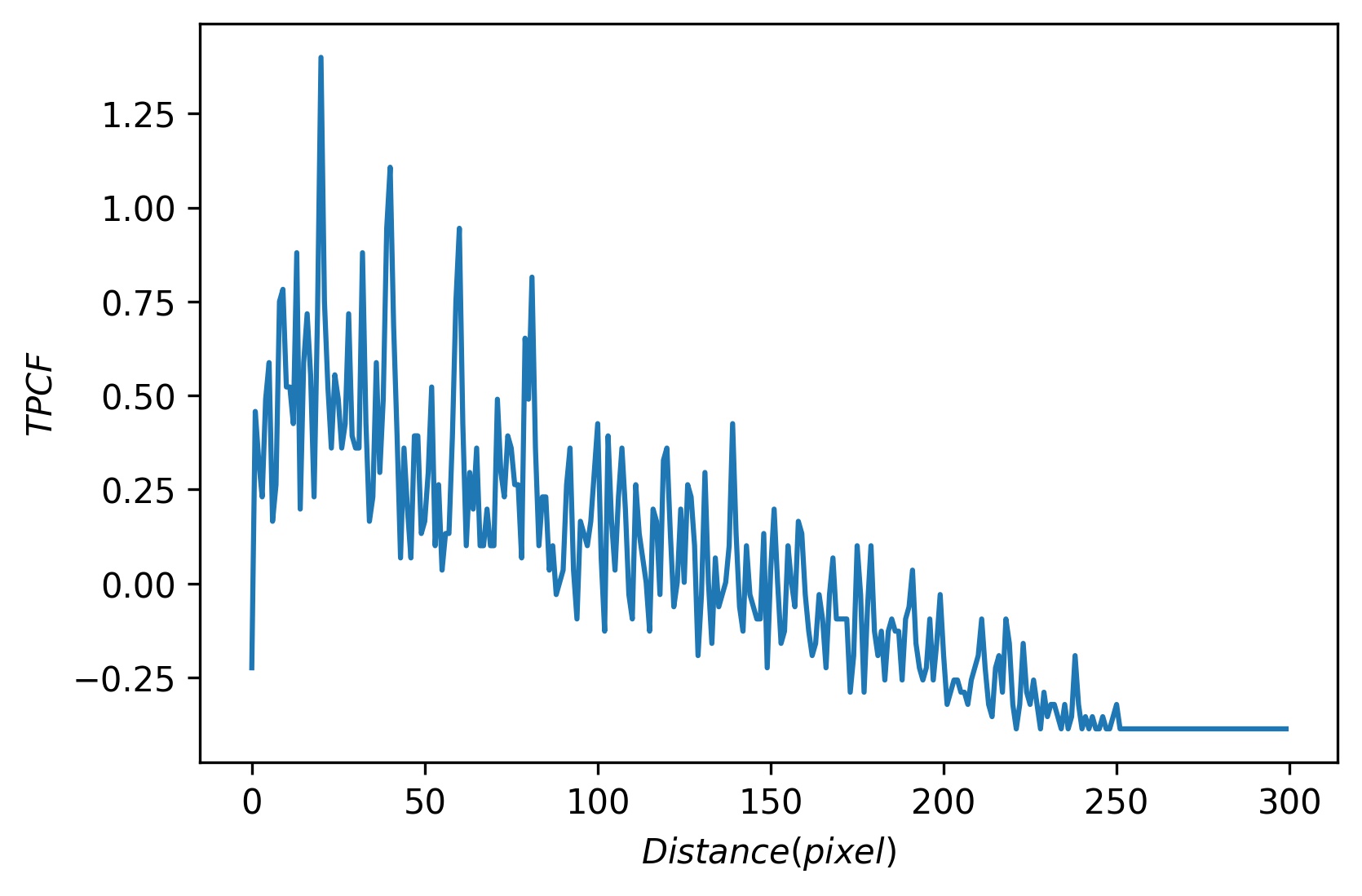}
    \end{subfigure}
    \caption{
    Two examples of the two-point correlation function computed on predicted ruler masks.
    Right: high-quality ruler prediction with clear peaks. Left: noisy prediction with degraded peaks.
    }
    \label{fig:tpcf}
\end{figure}

\paragraph{Pixel-scale estimation via the two-point correlation function.}
Let $\delta(\mathbf{x}) \in \{0,1\}$ be the binary ruler-mask indicator and $A$ the image area in pixels. The (isotropic) Two-Point Correlation Function (TPCF) \cite{jiao2007modeling,torquato2002random} is
\begin{equation}
    \xi_2(\Delta) = \frac{1}{A} \int d^2x \, \delta(\mathbf{x})\,\delta(\mathbf{x}+\Delta),
    \label{eq:tpcf}
\end{equation}
with $\Delta = \|\mathbf{x}_1-\mathbf{x}_2\|_2$ the Euclidean separation in pixels. When evaluated on ruler masks, $\xi_2(\Delta)$ exhibits peaks at separations corresponding to tick spacing (see \suppfig{fig:tpcf}). We compute $\xi_2$ on 300 discrete separations and feed the resulting 1D signature into a lightweight CNN regressor (three 1D-convolutional layers with max pooling and batch normalization, followed by a linear head) to predict $\rho$ = pixels-per-millimeter. Supervision for $\rho$ is obtained directly from the synthesis process by tracking the applied ruler transformations and known physical tick spacing.
\paragraph{Geometric features from the lesion contour.}
Let $M \in \{0,1\}^{H \times W}$ be the predicted lesion mask, let $\rho$ be the estimated pixels-per-millimeter, and let $\alpha = 1/\rho$ denote pixels-per-millimeter. From $M$ we extract one or more closed iso-contours at level $\nu$ (marching squares), each returned as an ordered sequence of sub-pixel points $\{(x_k,y_k)\}_{k=1}^{K}$ with $(x_{K+1},y_{K+1})\equiv(x_1,y_1)$~\cite{Mantz2008}. Using these vertices, we compute:

\emph{Perimeter (mm):} the Euclidean length of the polyline,
\begin{equation}
    P_\mathrm{lesion} \;=\; \alpha \sum_{k=1}^{K} \sqrt{(x_{k+1}-x_k)^2 + (y_{k+1}-y_k)^2},
    \label{eq:perimeter}
\end{equation}
which matches the implementation that sums consecutive point-to-point distances. (For strictly grid-adjacent steps this reduces to the familiar 8-connected chain-code weights $\{1,\sqrt{2}\}$~\cite{Dorst1987}.)

\emph{Area (mm$^2$):} the signed polygon area via Green’s theorem (shoelace formula),
\begin{equation}
    A_\mathrm{lesion} \;=\; \alpha^2 \,\frac{1}{2}\,\bigg|\sum_{k=1}^{K} \big(x_k\,y_{k+1} - y_k\,x_{k+1}\big)\bigg|,
    \label{eq:area}
\end{equation}
\textit{i.e.}, the absolute value of the oriented contour area scaled by $\alpha^2$. For multiple disjoint components, areas are summed; if interior holes are present, their (oppositely oriented) contour areas subtract naturally~\cite{ORourke1998}.

\emph{Radius of Gyration (mm):} measures the spatial dispersion of contour points about their centroid,
\begin{equation}
R_{\mathrm{g}}^2 = \frac{1}{N}\sum_{i=1}^N\big[(x_i-x_c)^2+(y_i-y_c)^2\big], \qquad
x_c=\frac{1}{N}\sum_{i=1}^N x_i,\; y_c=\frac{1}{N}\sum_{i=1}^N y_i,
\end{equation}
which is translation-invariant and, when normalized by size, provides a compact geometric descriptor complementary to area and perimeter for characterizing lesion extent, elongation, and boundary irregularity, often improving discrimination in wound classification and healing-monitoring tasks~\cite{hosseinabadi2012geometrical}.

\subsection{Graph Neural Network} \label{sec:gnn}
\begin{figure}[!b]
    \centering
    \includegraphics[width=\textwidth]{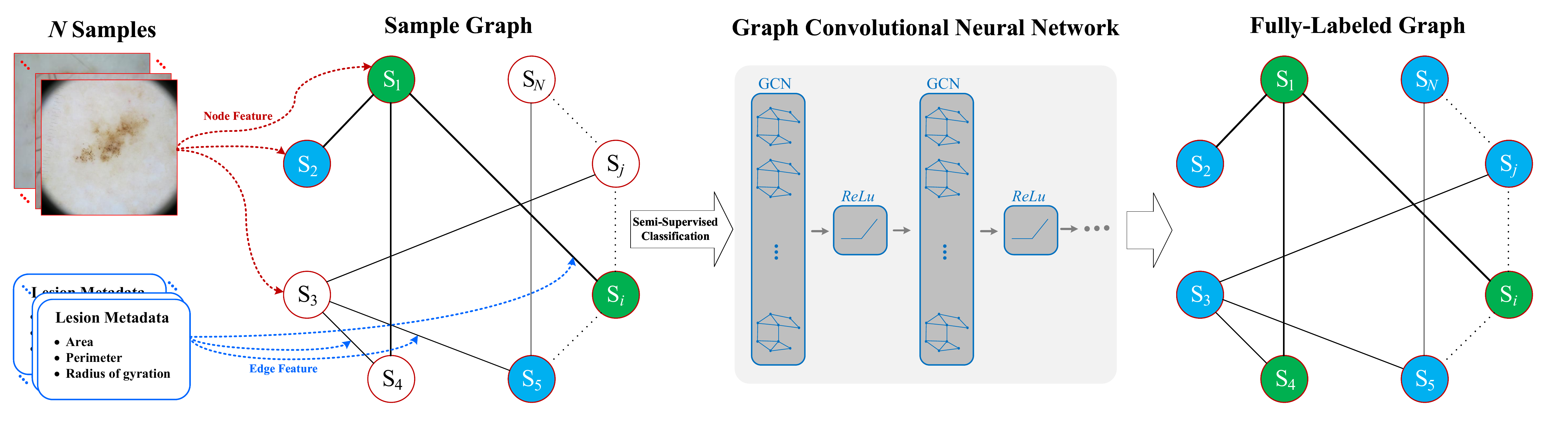}
    \caption{
        Overview of the proposed graph-based framework for classification using graph neural networks. Imaging data are used to construct node features, while metadata provide the basis for edge features. In the semi-supervised setting, only a subset of nodes have classification labels during training; after training, labels are inferred for all nodes.
    }
    \label{fig:pipeline}
\end{figure}

Inspired by \citep{parisot2018disease}, an overview of constructing the graph structure is shown in \autoref{fig:pipeline}.

For the graph structure, a population of $N$ samples $S_1, S_2, \dots, S_N$, each corresponding to a single dermoscopic image. Our goal is to predict the classification of each sample based on its imaging and auxiliary data.

The population sample can be denoted as a sparse weighted graph $G = (V, E, W)$, where $W$ is the graph adjacency matrix. Each dermoscopic image is denoted as a node (vertex) $v \in V$, corresponds to a sample $S_v$ of the population, and is accompanied by a $d$-dimensional feature vector $\mathbf{x}_v$ obtained from imaging data. The graph edges $E$ quantify the similarity between the samples by incorporating relevant auxiliary and imaging data.

By assigning a label $l \in \{ 0, 1, 2, \dots, 7 \}$ to each graph node to indicate a normal label or the presence of any of the aforementioned skin abnormalities, we model skin lesion prediction as a node classification problem. For training the model, we assume a semi-supervised learning scheme, in which the graph neural network receives the complete population graph along with all the node and edge features and only an initial subset of the graphs nodes are used during the training and optimization process.

Intuitively, the population graph acts as a regulator, encouraging vertices connected with high weights to help filter their neighbors features for maximizing label propagation. Therefore, constructing the population graph is a key aspect of the method, as an ill-structured population graph (that does not accurately describe the similarity between samples and their feature vectors) will fail to exploit the power of the graph neural networks. Moreover, an improperly structured population graph can even perform worse compared to a linear classifier. Similarly, this equates to performing an image convolution operation on unrelated pixels (\textit{e.g.} randomly selected) instead of a local image patch.

Constructing the population graph structure requires two primary decisions;

\begin{enumerate}
    \item The definition of the feature vector $\mathbf{x}_v$ that describes each node,
    \item The edges of the graph $E$ and their weights $W$, which describe the similarity between the vertices and their feature vectors.
\end{enumerate}

As the objective of this research was to study the improvement on image classification performance when additional information is used in companion with the imaging data, we will only utilize the imaging data of a node's corresponding sample for deriving its feature vector.

We employ two different methods. In the first method, the model's initial weights are pre-trained only on the \textit{ImageNet} \citep{deng2009imagenet} dataset. However in the second method, in addition to \textit{ImageNet} transfer-learning, we also train the model on the 19{,}627 samples mentioned in \autoref{sec:dataset}.
In both methods, the imaging data with dimensions $300\times300$ and basic ImageNet normalization is fed to the feature-extractor block of an EfficientNet-B3 model \citep{tan2019efficientnet} and the output vector of length 1536 is used as the feature vector of the node corresponding to the image's sample.
A breakdown on the number of samples used for the training and validation sets used for training the feature-extractor model can be found in \supptab{table:isic2019-trainval}.

Similar to the pixel neighborhood logic used by the convolutional filter in CNNs, the graph structure provides a wider field of view for label propagation and filters the value of a feature with respect to its neighbors instead of examining each feature separately. Thus, for an accurate modeling of interactions between the nodes feature vectors, a carefully designed graph edge structure is required.
Our hypothesis is that non-image auxiliary data can provide key information to explain the relationship between the feature vectors of the samples.
The purpose of using this information is to define an accurate neighborhood system (similar to the convolution filter in the image) that improves the performance of the convolution operation in the graph structure.

Four methods of deriving edge features are explored in this research; fully-weighted, threshold, random, and identical. In all four methods, the graph structure is assumed fully-connected and each edge is then weighted accordingly.

In the fully-weighted method, inspired by \cite{parisot2018disease}, the weight of each edge is determined by applying a similarity function on the auxiliary data of the node samples at the two ends of the edge.
Given a set of $H$ non-imaging auxiliary metrics $M = \{ m_h | h \in [1, H]\}$ (\textit{e.g.} gender or age of the sample), the adjacency matrix of the population graph $w$ is defined as:

\begin{equation}
    W(v, w) = Sim(\mathbf{x}_v, \mathbf{x}_w) \sum_{h = 1}^{H} \mathbf{w}_h . \gamma(m_h(v), m_h(w)),
    \label{eq:parisot}
\end{equation}

where $\mathbf{w}_h$ is the impact weight of the auxiliary metric $m_h$, $\gamma$ is an operator measuring the distance between auxiliary metrics, and $Sim(\mathbf{x}_v, \mathbf{x}_w)$ is a measure of similarity between the feature vector of vertices, which increases the weight of edges between the most similar vertices of the graph.

The function to calculate the similarity of node feature vectors $Sim(\mathbf{x}_v, \mathbf{x}_w)$ is an integral component in the construction of \autoref{eq:parisot} as seen in \cite{parisot2018disease}. However in this research, we omit this function from the edge weight definition.
As previously discussed, the samples' auxiliary data is obtained from training a naturally error-prone model on synthesized data resulted from other error-prone methods. Even if the final result is accurate for independent classification and segmentation problems, the resulting sample feature vector can introduce unwanted and unexpected errors in the diffusion mechanism of the graph structure.

Function $\gamma$ is defined differently depending on the type of the auxiliary data metric. Each sample in the dataset is associated with three named metrics; gender, anatomical site of the dermoscopy image on the body, name of the source dataset, and four numerical metrics; age, lesion area, lesion perimeter, and lesion radius of gyration.

For named metrics, $\gamma$ is defined as a Kronecker delta function with the following formula:

\begin{equation}
    \gamma(m_h(v), m_h(w)) =
    \begin{cases}
        1 & \text{if } m_h(v) = m_h(w), \\
        0 & \text{otherwise}.
    \end{cases}
\end{equation}

The definition of $\gamma$ is however slightly more contrived for numerical metrics. For each of the four numerical metric, the following procedure is applied:

First, a cumulative distribution function is calculated for all the metric values of the entire sample population. To deal with outliers, values below the $1^{\text{st}}$ percentile and above the $99^{\text{th}}$ percentile are clipped to the respective percentile values. Next, for each node pair $v, w \in V$, the set $F = \{ f_{vw} \mid f_vw = |m_h(v) - m_h(w)| \}$ is calculated and a $z$-score normalization is performed to bring the mean and variance values to 0 and 1.

By denoting the newly normalized set as $F^\prime$, the function $\gamma$ for numerical metric $m_h$ is defined as the following:

\begin{equation}
    \gamma(m_h(v), m_h(w)) = Norm(f^\prime_{vw})
\label{eq:norm}
\end{equation}

for which, $Norm(x) = -\tanh(x)$ was selected which results in the value of $\gamma$ consequently falling in the $[-1, 1]$ range.

For the threshold method of deriving edge features, the edge weight is calculated is as

\begin{equation}
    W(v, w) =
    \begin{cases}
        1 & \text{if } W(v, w) \geq T, \\
        0 & \text{otherwise}.
    \end{cases}
\end{equation}

in which $W(v, w)$ is the weight calculated by \autoref{eq:parisot} using operator $\gamma$ from \autoref{eq:norm} and $T$ is a manually selected threshold value.

In the random method, each edge receives a random weight value in the range $[-1, 1]$ drawn from a uniform distribution. Then, from the set of edged sorted ascending by their weight value, the top $n$ edges are selected, where $n$ is the number of edges that have a non-zero value in the threshold method that had the best performance in \autoref{tab:thresholds}.

Finally in the identical method, all the edge weights have the same value of 1.

For \emph{GraphDerm}, the graph neural network architecture involves 32 graph convolution hidden layers derived from the spectral-based definition of graph convolution in \cite{kipf2016semi} with a Rectified Linear Unit (ReLU) activation and a 0.5 dropout rate.

The graph neural network model is trained on the whole of the population graph as its input. The training set, following the semi-supervised learning scheme, involves a labeled subset of vertices at the beginning and the validation subset (the unlabeled vertices) receive their features during the training process.

The model was trained with a batch size of 256 and using Adam optimizer \cite{kingma2014adam} with an initial learning rate of 1\% for 300 epochs, which stops training if the performance does not improve for 50 consecutive epochs.

A 5-fold cross-validation scheme and also a weighted cross-entropy loss function are used for the optimization process. The weight of the loss function for the presence ($+$) or the lack there of ($-$) a class $i$ is calculated using

\begin{equation}
    w^{+/-}_i = \frac{\# \text{classes all in samples all }}{\# \text{ } {+/-} \text{ class in sample } i}
\end{equation}

and the final value of the loss function is obtained using the following:

\begin{equation}
    loss = \frac{1}{8} \sum_{i=0}^{7} w^{+/-}_iloss_i .
    \label{eq:loss}
\end{equation}

After training the graph neural network model, The raw predicted values passed through a softmax function are calculated on the test set, and labels are assigned to the unlabeled vertices accordingly.
For the final model, general details of the utilized parameters and the number of samples in each class alongside their corresponding loss function weights can be found in \supptab{table:parameters} and \supptab{table:weights}, respectively.

All the methodology in this research have been implemented using the Python programming language version 3.7 \cite{van1995python}. The neural network models have been developed using Tensorflow version 2.7 \cite{tensorflow2015-whitepaper}, Keras version 2.7 \cite{chollet2015keras}, and have been trained using an Nvidia GeForce RTX 5000. The NetworkX package \cite{SciPyProceedings_11} was utilized for processing and analyzing the graphs, the scikit-learn package \cite{scikit-learn} for statistical analysis metrics mentioned in \cite{van2014scikit}, and finally the Matplotlib package was used for generating the plots.

\section{Results} \label{sec:results}
This section reports the empirical evaluation of the constituent modules and the proposed population-graph classifier. We follow the referencing style established earlier for figures, tables, and citations.

\subsection{Evaluation Protocol}\label{section:statistics}
To adopt a conservative and reproducible assessment, we report precision, recall, and the area under the ROC curve (AUC) for all experiments, together with confusion matrices when appropriate. Each underlying class is treated as a binary label (present/absent). Using a Bayesian formulation, we compute 95\% marginal credible intervals for all summary metrics, and we assess statistical significance via $p$-values, adopting a conservative $3\sigma$ decision level. Because ground-truth labels are binary whereas model outputs are probabilistic, a single operating point is selected to favor high specificity; ROC curves are provided to contextualize this choice.

\subsection{Synthesis Pipeline Performance}\label{section:data-results}
\begin{figure}[!h]
    \centering
    \begin{subfigure}{0.45\textwidth}
        \centering
        \includegraphics[width=\textwidth]{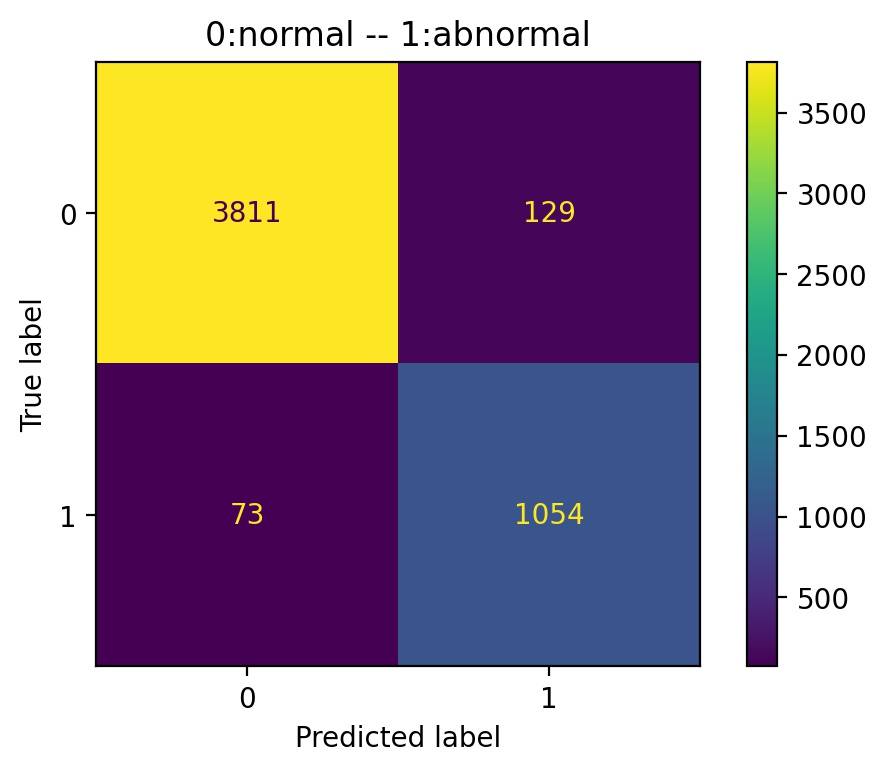}
    \end{subfigure}
    \hfill
    \begin{subfigure}{0.53\textwidth}
        \centering
        \includegraphics[width=\textwidth]{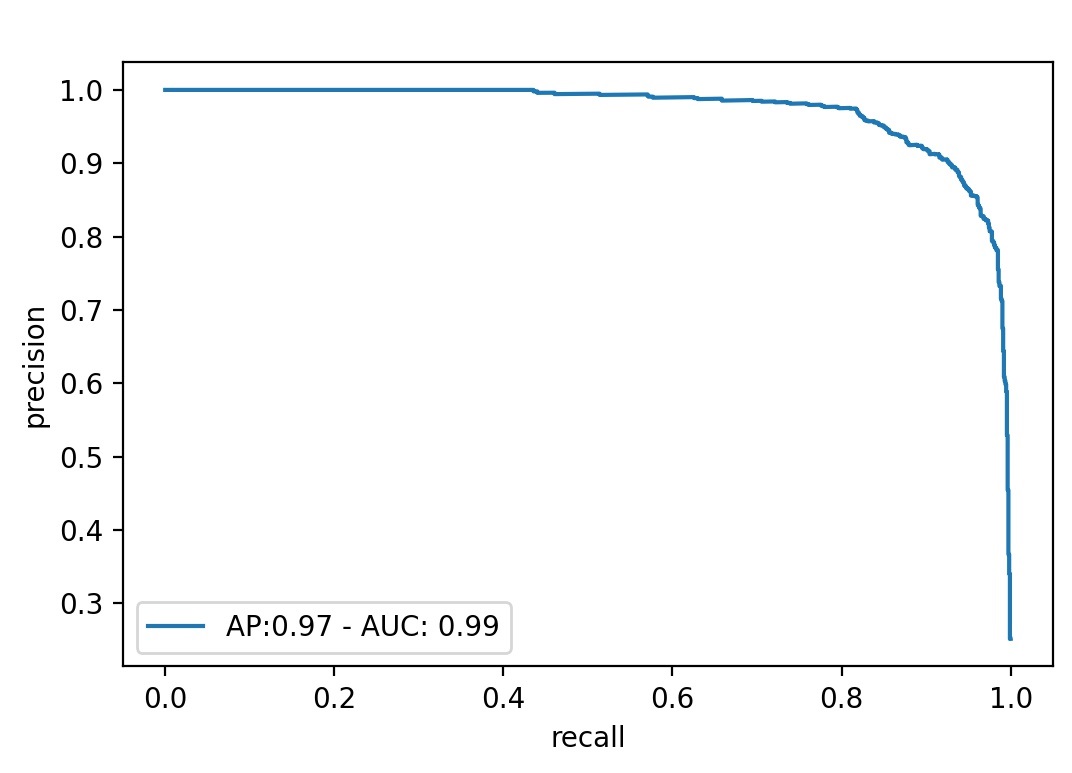}
    \end{subfigure}
    \caption{
    Inspection model for detecting embedded rulers: confusion matrix (left) and ROC curve (right).
    }
    \label{fig:seg-confusion-roc}
\end{figure}

The data-synthesis process begins with automated screening of ISIC 2019 images for the presence of embedded rulers using an EfficientNet-B2 classifier. The classifier achieves excellent discrimination with an AUC of $0.99$ (see confusion matrix and ROC in \suppfig{fig:seg-confusion-roc}).

\begin{figure}[!b]
    \centering
    \includegraphics[width=\textwidth]{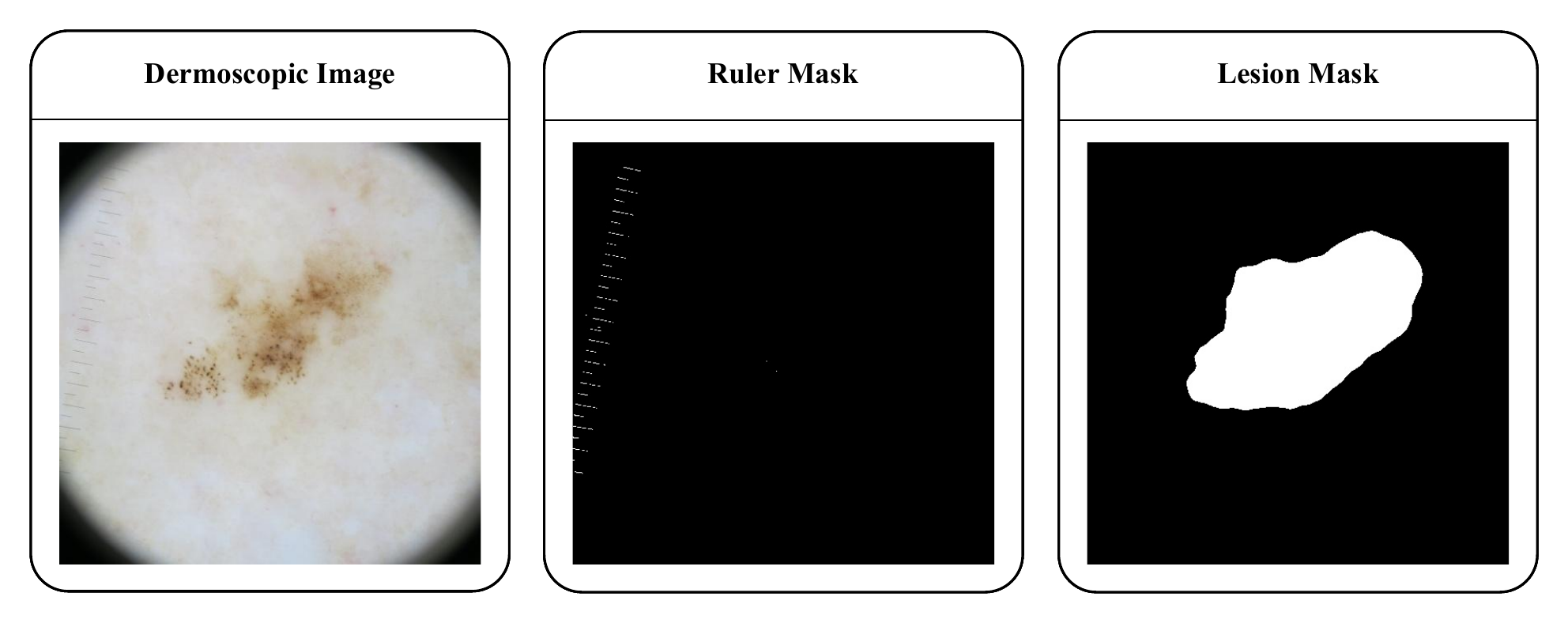}
    \caption{
        Example dermoscopy image with predicted ruler mask and predicted lesion mask.
    }
    \label{fig:sample}
\end{figure}

Next, two U-Net models were trained for ruler and lesion segmentation. Backbones based on MobileNetV2 and SE-ResNet-18 were compared. As summarized in \autoref{tab:seg-results}, SE-ResNet-18 offers superior lesion performance and comparable ruler performance at much lower wall-clock time for the $512 \times 512$ input setting; we therefore adopt SE-ResNet-18 for both tasks. A qualitative example is shown in \autoref{fig:sample}.

\begin{table}[!t]
    \centering
    \caption{
        The performance results of the two models for the problems of skin lesion segmentation and ruler segmentation with two different backbones.
    }
    \begin{tabular}{rrr}
    \toprule
    Problem                                 & Backbone        & Dice Score       \\ \midrule
    \multirow{2}{*}{Lesion Segmentation} & MobileNetV2  & $0.906 \pm 0.21$ \\
                                          & SE-ResNet-18 & $0.908 \pm 0.19$ \\ \hline
    \multirow{2}{*}{Ruler Segmentation}       & MobileNetV2  & $0.900 \pm 0.29$ \\
                                          & SE-ResNet-18 & $0.904 \pm 0.22$ \\
    \bottomrule
    \end{tabular}
    \label{tab:seg-results}
\end{table}

\begin{figure}[!h]
    \centering
    \begin{subfigure}{0.49\textwidth}
        \centering
        \includegraphics[width=\textwidth]{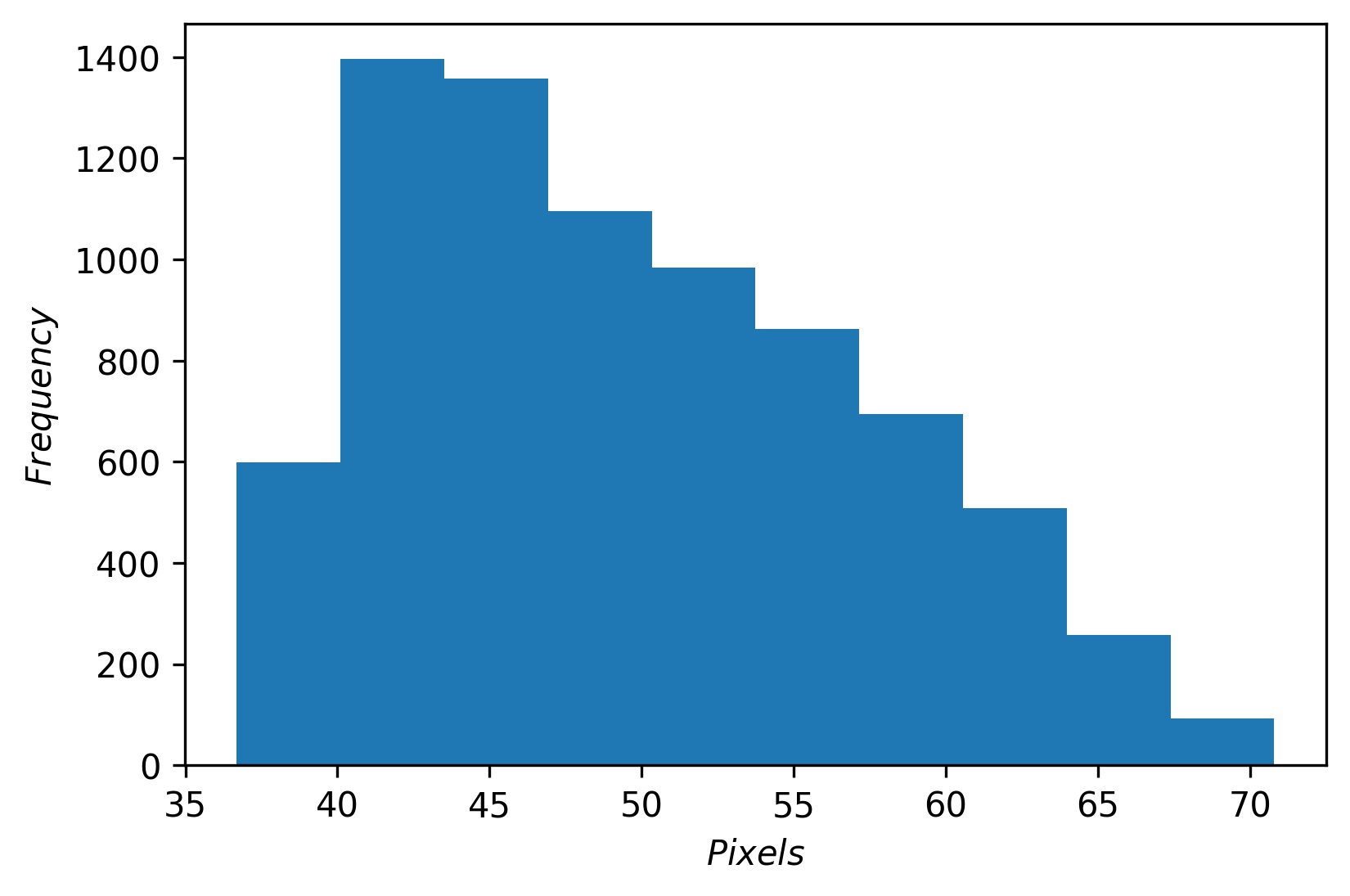}
    \end{subfigure}
    \hfill
    \begin{subfigure}{0.49\textwidth}
        \centering
        \includegraphics[width=\textwidth]{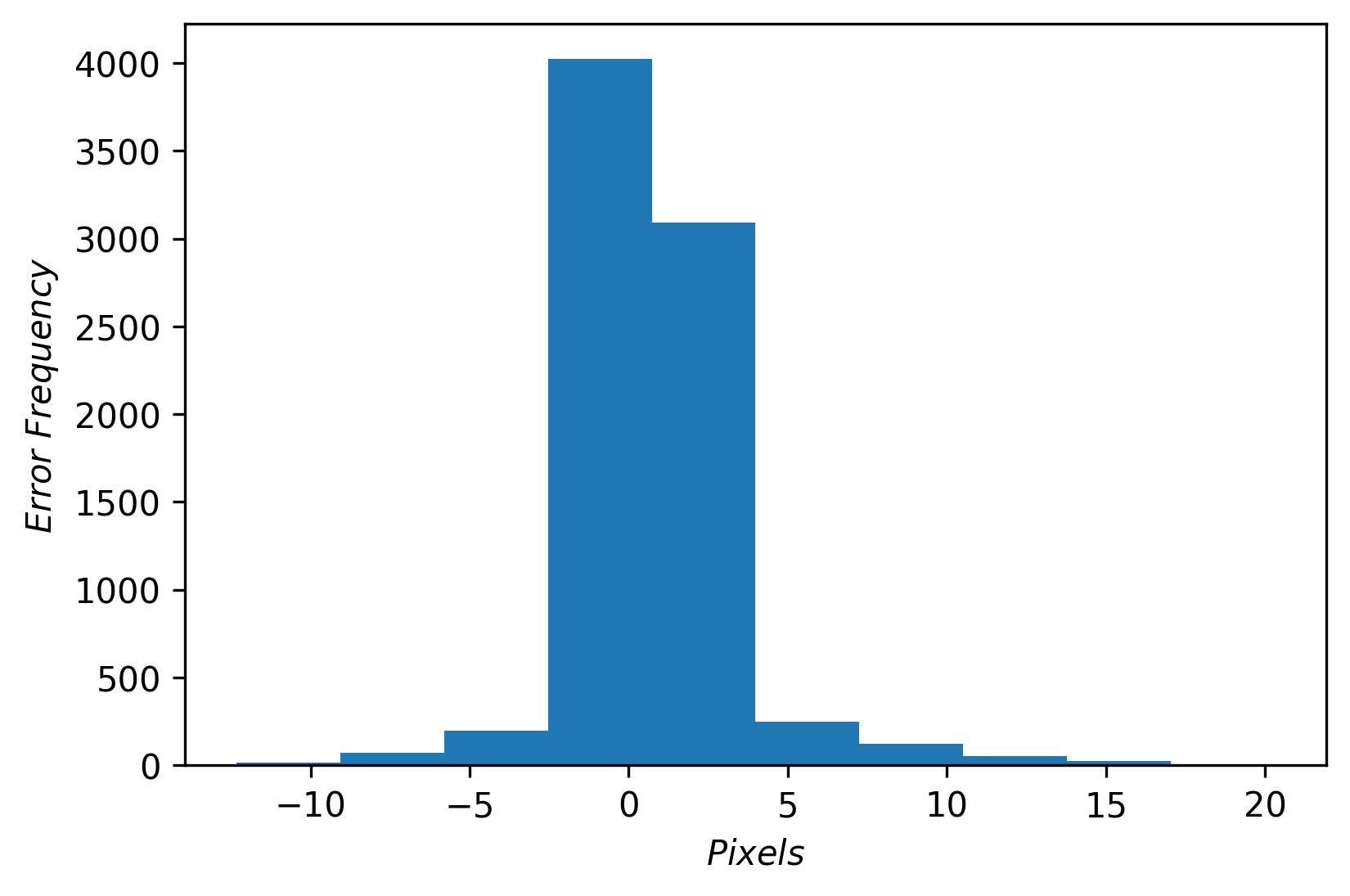}
    \end{subfigure}
    \caption{
    Pixel-scale regressor outputs across ruler types: model predictions (left) and prediction error histogram (right).
    }
    \label{fig:histograms}
\end{figure}

Finally, pixel scale (pixels-per-millimeter) is inferred by regressing on the two-point correlation signature computed from the predicted ruler masks. The regressor attains an RMSE of $6.6$ pixels and an MAE of $1.5$ pixels. Distributional summaries across ruler types are shown in \suppfig{fig:histograms}.

\subsection{Population-Graph Classification}\label{section:results}
We evaluate the proposed graph neural network on ISIC 2019 under the semi-supervised setting described previously, comparing two node-feature initializations (ImageNet vs.\ additional fine-tuning on ISIC 2019 imaging data and four edge-weighting schemes (full-weighted, thresholded, random, identical. The main results are compiled in \autoref{tab:results}.

\begin{itemize}
    \item \textbf{Node features.} Using features initialized and further trained on ISIC 2019 yields large gains over ImageNet-only for both the ANN baseline and all GNN variants.
    \item \textbf{Edge construction.} The full-weighted strategy delivers the best overall performance (highest AUC and recall, with strong precision). The identical and random baselines underperform, underscoring the importance of a meaningful neighborhood system.
\end{itemize}

\begin{table}[!t]
    \centering
    \caption{
        Combined results for node-feature and edge-weighting strategies (number of active edges shown per setting). Best values per metric are highlighted in blue.
    }
    \small
    \begin{tabular}{llllll}
    \toprule
    Method                           & Initialization & \# Edges     & Precision (95\% CI)   & Recall (95\% CI)      & AUC (95\% CI)       \\
    \midrule
    \multirow{2}{*}{ANN}             & ImageNet       & -           & 0.8372 $\pm$ 0.0791   & 0.5211 $\pm$ 0.2928   & 0.8701 $\pm$ 0.0421 \\
                                     & ISIC 2019      & -           & 0.7416 $\pm$ 0.0229   & 0.7173 $\pm$ 0.0168   & 0.9440 $\pm$ 0.0064 \\
    \hline
    \multirow{2}{*}{Full-Weight}     & ImageNet       & 2344695     & \textcolor{blue}{0.8524 $\pm$ 0.0769}   & 0.5307 $\pm$ 0.3075   & 0.8773 $\pm$ 0.0903 \\
                                     & ISIC 2019      & 2344695     & 0.8429 $\pm$ 0.0190   & \textcolor{blue}{0.8652 $\pm$ 0.0185}   & \textcolor{blue}{0.9812 $\pm$ 0.0020} \\
    \hline
    \multirow{2}{*}{Threshold}       & ImageNet       & 36651       & 0.7880 $\pm$ 0.1304   & 0.6180 $\pm$ 0.1395   & 0.9098 $\pm$ 0.0534 \\
                                     & ISIC 2019      & 36651       & 0.8321 $\pm$ 0.0159   & 0.8490 $\pm$ 0.0110   & 0.9788 $\pm$ 0.0019 \\
    \hline
    Random                           & -              & 36651       & 0.7001 $\pm$ 0.0220   & 0.7067 $\pm$ 0.0209   & 0.9330 $\pm$ 0.0106 \\
    \hline
    Identical                        & -              & 2344695     & 0.6426 $\pm$ 0.0420   & 0.6047 $\pm$ 0.0640   & 0.9184 $\pm$ 0.0119 \\
    \bottomrule
    \end{tabular}
    \label{tab:results}
\end{table}

Training dynamics (loss, precision, recall, and AUC for both training and validation) for the full-weighted and thresholded graphs are provided in \suppfig{fig:epochs-weighted} and \suppfig{fig:epochs}, respectively.

\begin{figure}[!h]
    \centering
    \begin{subfigure}{\textwidth}
        \centering
        \includegraphics[width=\textwidth]{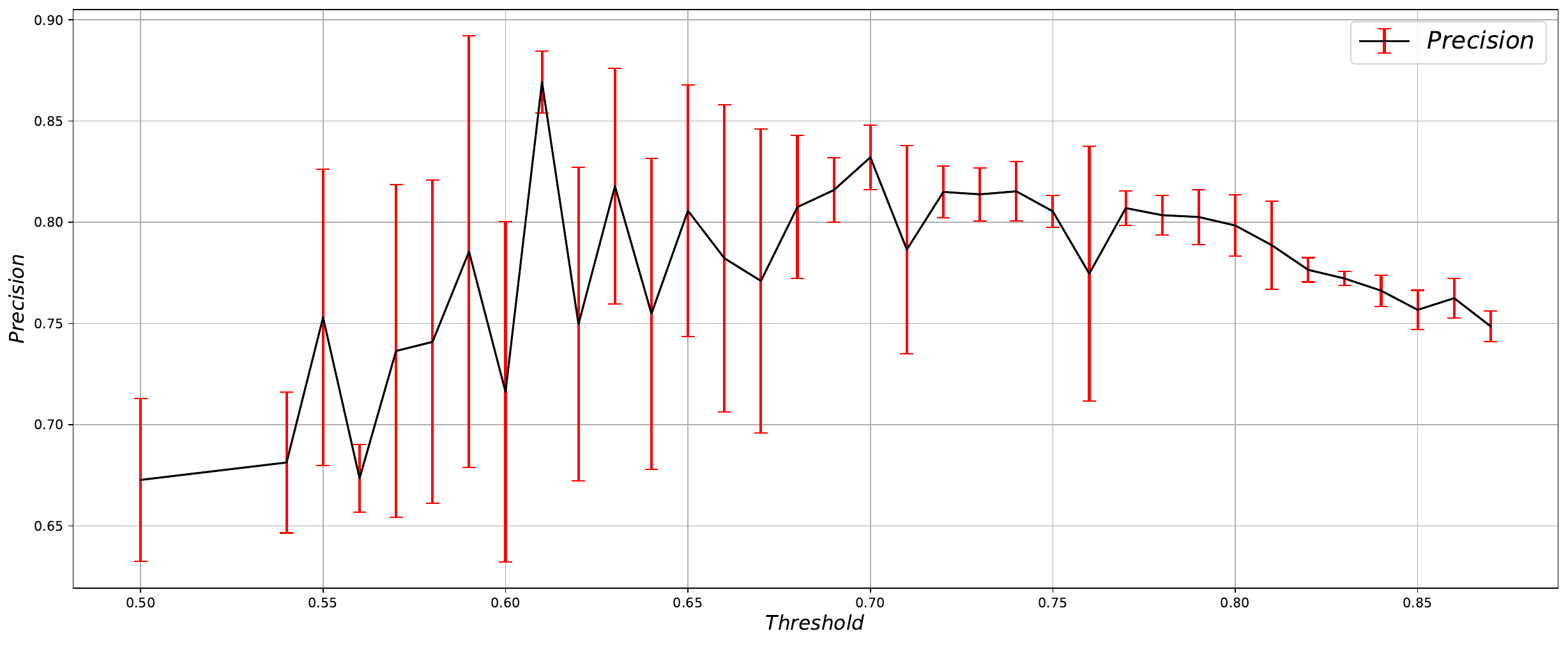}
    \end{subfigure}
    \hfill
    \begin{subfigure}{\textwidth}
        \centering
        \includegraphics[width=\textwidth]{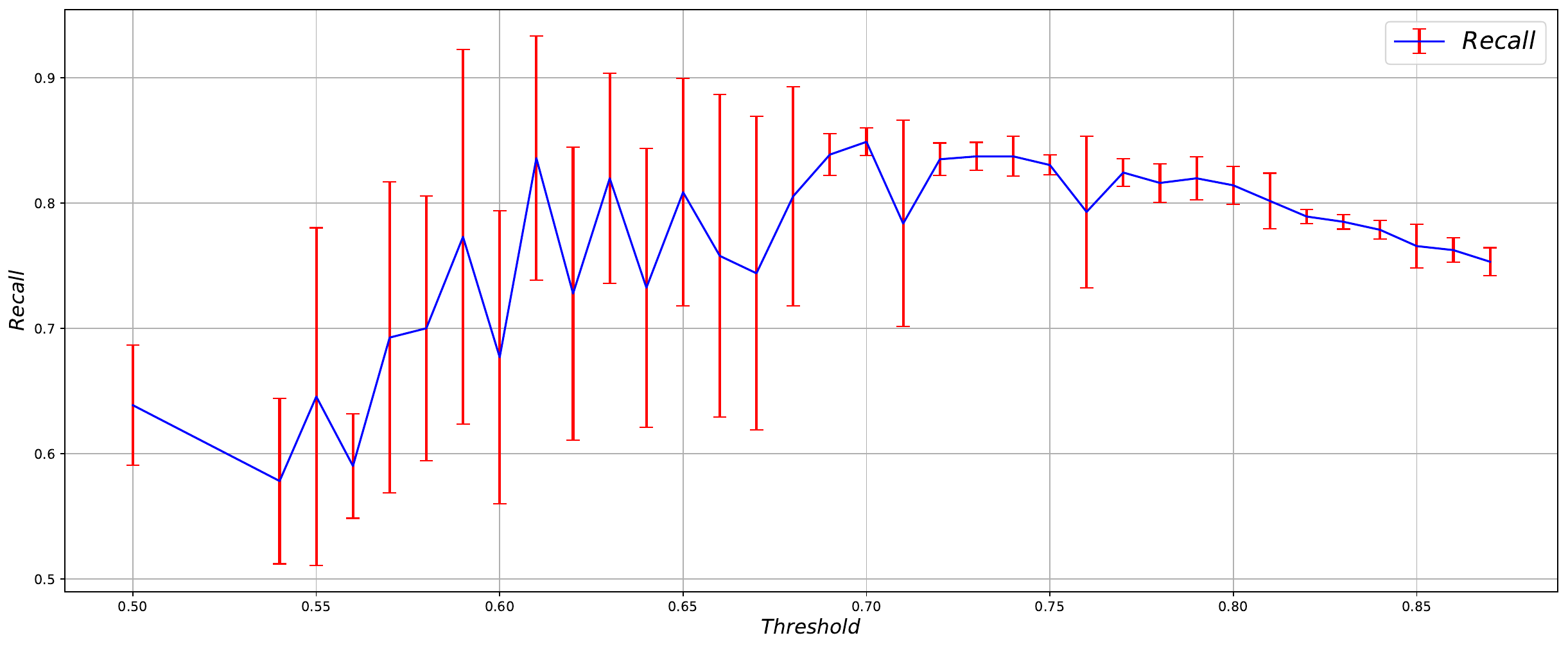}
    \end{subfigure}
    \hfill
    \begin{subfigure}{\textwidth}
        \centering
        \includegraphics[width=\textwidth]{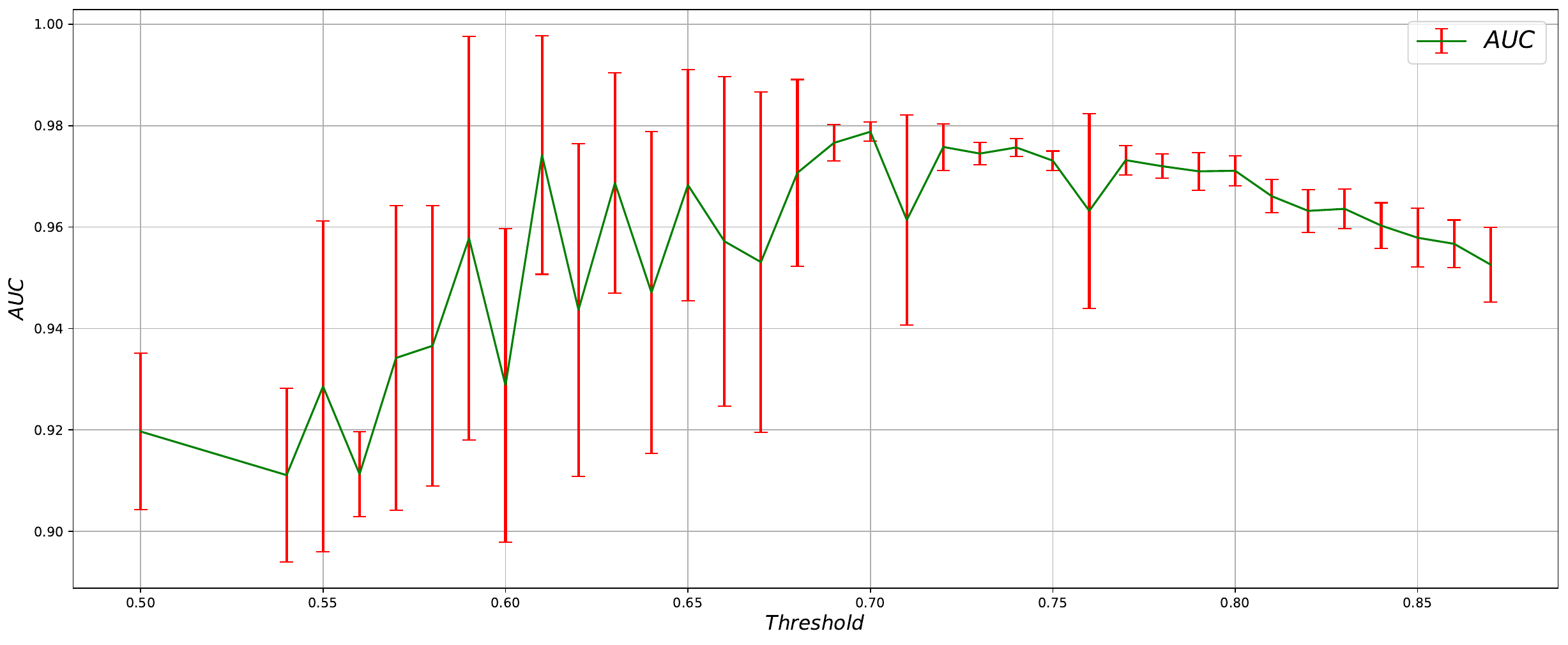}
    \end{subfigure}
    \caption{
        Precision, Recall, and AUC (with 95\% error bars) for the thresholded edges across thresholds $T \in [0,1]$.
    }
    \label{fig:metric-thresholds}
\end{figure}

\begin{figure}[!h]
    \centering
    \begin{subfigure}{0.9\textwidth}
        \centering
        \includegraphics[width=\textwidth]{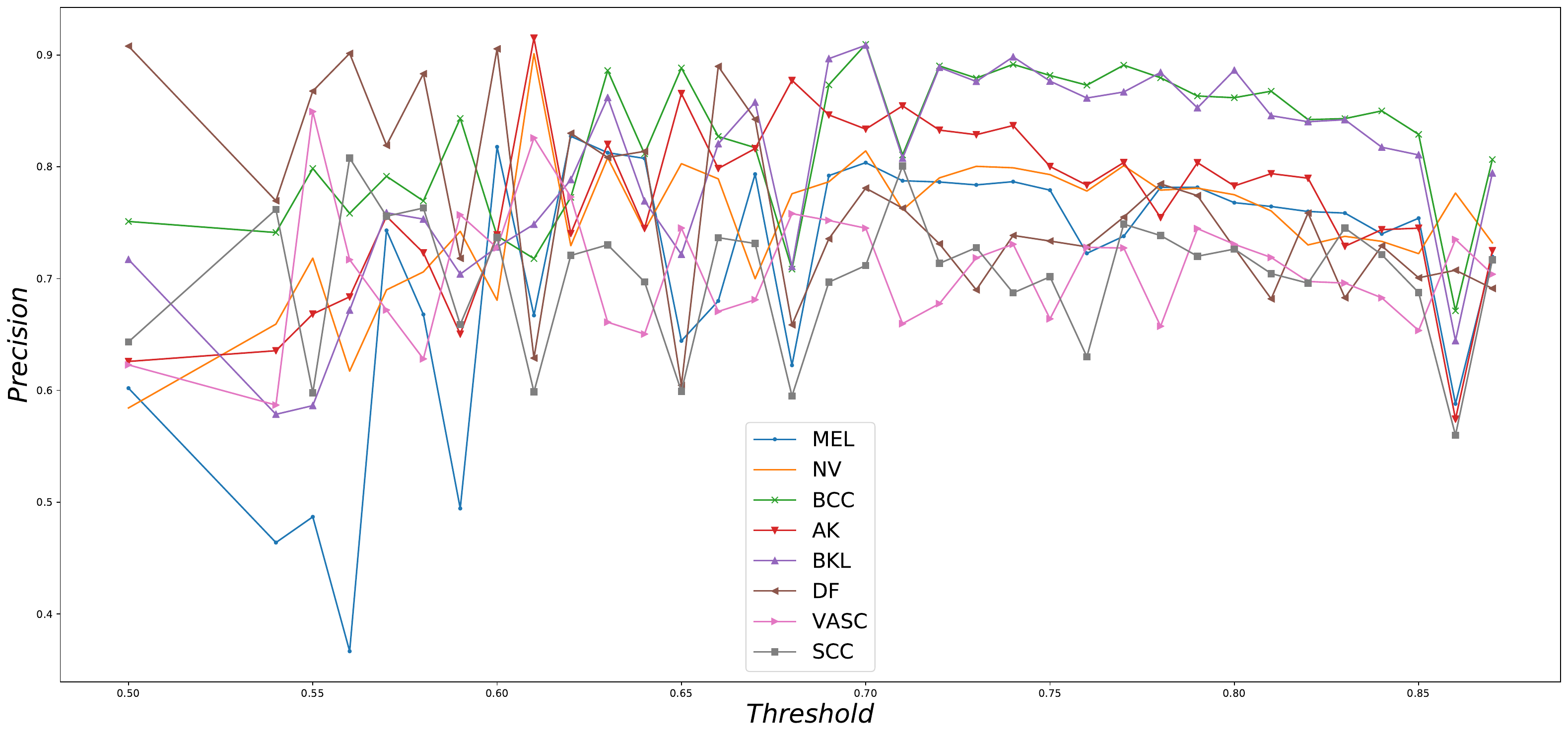}
    \end{subfigure}
    \hfill
    \begin{subfigure}{0.9\textwidth}
        \centering
        \includegraphics[width=\textwidth]{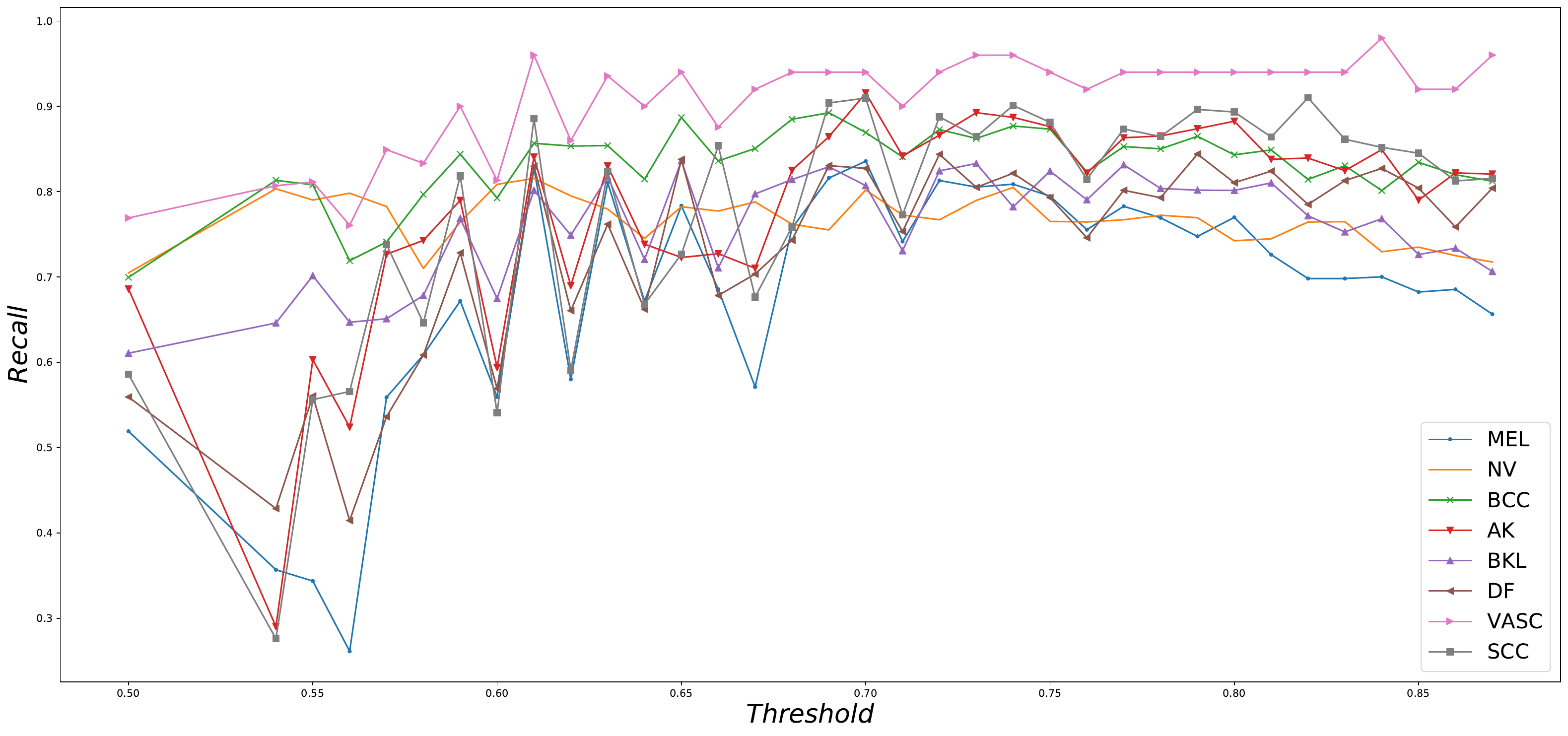}
    \end{subfigure}
    \hfill
    \begin{subfigure}{0.9\textwidth}
        \centering
        \includegraphics[width=\textwidth]{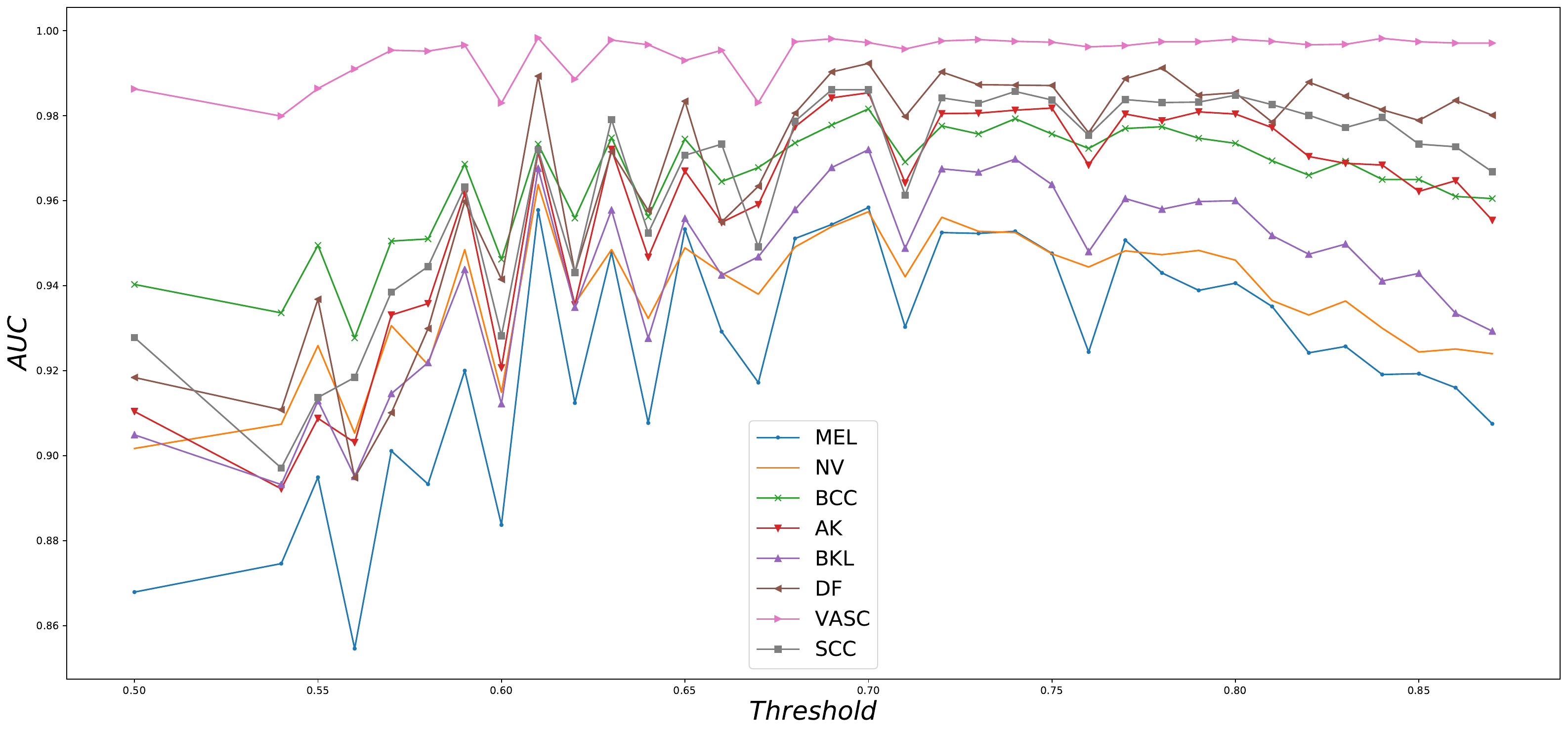}
    \end{subfigure}
    \caption{
    Performance chart of the threshold edge feature extraction method (with the ISIC 2019 Initialization-based node feature extraction method in the statistical measures of precision, recall, and area under the receiver operating characteristic curve for different threshold values, separated by dataset classifications.
    }
    \label{fig:class-thresholds}
\end{figure}

To probe sparsification, we sweep the threshold $T \in [0,1]$ in increments of $0.05$ and record performance and the resulting number of edges. As shown in \autoref{tab:thresholds}, $T{=}0.7$ yields the best overall trade-off across all metrics; detailed metric trends with 95\% error bars are plotted in \autoref{fig:metric-thresholds}. \textit{Class-wise} precision, recall, and AUC as functions of $T$ are provided in \suppfig{fig:class-thresholds}.

\begin{table}[!b]
    \centering
    \caption{
        Performance of the threshold edge feature deriving method with node feature extraction method based on ISIC 2019 initialization for different thresholds. The best threshold performance for each criterion is highlighted in blue.
    }
    \begin{tabular}{lllll}
    \toprule
    Threshold   & \# Edges  & Precision (95\% CI)    & Recall (95\% CI)       & AUC (95\% CI)      \\
    \midrule
    0.0         & 1045454   & 0.6808 $\pm$ 0.0345    & 0.6102 $\pm$ 0.0613    & 0.9199 $\pm$ 0.0132 \\
    0.5         & 158138    & 0.6727 $\pm$ 0.0401    & 0.6388 $\pm$ 0.0478    & 0.9197 $\pm$ 0.0154 \\
    0.55        & 110307    & 0.7530 $\pm$ 0.0731    & 0.6457 $\pm$ 0.1347    & 0.9286 $\pm$ 0.0326 \\
    0.6         & 66647     & 0.7162 $\pm$ 0.0840    & 0.6771 $\pm$ 0.1171    & 0.9288 $\pm$ 0.0309 \\
    0.65        & 47054     & 0.8057 $\pm$ 0.0622    & 0.8088 $\pm$ 0.0908    & 0.9683 $\pm$ 0.0228 \\
    \textcolor{blue}{0.7}         & \textcolor{blue}{36651}     & \textcolor{blue}{0.8321 $\pm$ 0.0159}    & \textcolor{blue}{0.8490 $\pm$ 0.0110}    & \textcolor{blue}{0.9788 $\pm$ 0.0019} \\
    0.75        & 26957     & 0.8054 $\pm$ 0.0079    & 0.8305 $\pm$ 0.0079    & 0.9731 $\pm$ 0.0019 \\
    0.8         & 17493     & 0.7984 $\pm$ 0.0151    & 0.8143 $\pm$ 0.0152    & 0.9711 $\pm$ 0.0030 \\
    0.85        & 7959      & 0.7567 $\pm$ 0.0097    & 0.7658 $\pm$ 0.0174    & 0.9579 $\pm$ 0.0058 \\
    \bottomrule
    \end{tabular}
    \label{tab:thresholds}
\end{table}

Finally, per-class analyses for the best full-weighted configuration and for the thresholded graph at $T{=}0.7$ are reported in \autoref{tab:t3} and \autoref{tab:t2}. Both strategies perform particularly well for \emph{benign keratosis} and \emph{vascular lesion} across metrics.

\begin{table}[!h]
    \centering
    \caption{
        Per-class precision, recall, and AUC for the full-weighted edges with ISIC 2019–initialized node features.
    }
    \small
    \begin{tabular}{llll}
        \toprule
        Classes                 & Precision (95\% CI) & Recall (95\% CI)   & AUC (95\% CI)      \\
        \midrule
        Melanoma                & 0.8723 $\pm$ 0.0305  & 0.8464 $\pm$ 0.0263  & 0.9754 $\pm$ 0.0033 \\
        Melanocytic Nevus       & 0.8794 $\pm$ 0.0317  & 0.8806 $\pm$ 0.0322  & 0.9837 $\pm$ 0.0037 \\
        Basal Cell Carcinoma    & 0.8668 $\pm$ 0.0197  & 0.8942 $\pm$ 0.0280  & 0.9837 $\pm$ 0.0016 \\
        Actinic Keratosis       & 0.8040 $\pm$ 0.0497  & 0.8252 $\pm$ 0.0334  & 0.9746 $\pm$ 0.0044 \\
        Benign Keratosis        & \textcolor{blue}{0.9033 $\pm$ 0.0172}  & 0.8910 $\pm$ 0.0166  & 0.9863 $\pm$ 0.0025  \\
        Dermatofibroma          & 0.6778 $\pm$ 0.0393  & 0.7473 $\pm$ 0.0841  & 0.9813 $\pm$ 0.0031  \\
        Vascular Lesion         & 0.8400 $\pm$ 0.0655  & \textcolor{blue}{0.9000 $\pm$ 0.0640}  & \textcolor{blue}{0.9891 $\pm$ 0.0007} \\
        Squamous Cell Carcinoma & 0.6960 $\pm$ 0.0363  & 0.8383 $\pm$ 0.0695  & 0.9754 $\pm$ 0.0051 \\
        \midrule
        Average                 & 0.8174 $\pm$ 0.0362  & 0.8529 $\pm$ 0.0443  & 0.9812 $\pm$ 0.0031 \\
        \bottomrule
    \end{tabular}
    \label{tab:t3}
\end{table}

\begin{table}[!h]
    \centering
    \caption{
        Per-class precision, recall, and AUC for the \emph{thresholded} edges at $T{=}0.7$ with ISIC 2019–initialized node features.
    }
    \small
    \begin{tabular}{llll}
        \toprule
        Classes                 & Precision (95\% CI) & Recall (95\% CI)   & AUC (95\% CI)      \\
        \midrule
        Melanoma                & 0.8037 $\pm$ 0.0178  & 0.8357 $\pm$ 0.0454 & 0.9584 $\pm$ 0.0047 \\
        Melanocytic Nevus       & 0.8142 $\pm$ 0.0569  & 0.8020 $\pm$ 0.0339  & 0.9574 $\pm$ 0.0088 \\
        Basal Cell Carcinoma    & 0.9095 $\pm$ 0.0195  & 0.8694 $\pm$ 0.0323 & 0.9816 $\pm$ 0.0034 \\
        Actinic Keratosis       & 0.8337 $\pm$ 0.0557  & 0.9156 $\pm$ 0.0229 & 0.9854 $\pm$ 0.0065 \\
        Benign Keratosis        & \textcolor{blue}{0.9087 $\pm$ 0.0439}  & 0.8071 $\pm$ 0.0191 & 0.9720 $\pm$ 0.0121  \\
        Dermatofibroma          & 0.7809 $\pm$ 0.0599  & 0.8273 $\pm$ 0.0661 & 0.9923 $\pm$ 0.0040  \\
        Vascular Lesion         & 0.7451 $\pm$ 0.0937  & \textcolor{blue}{0.9400 $\pm$ 0.1200}     & 0.9972 $\pm$ 0.0029 \\
        Squamous Cell Carcinoma & 0.7117 $\pm$ 0.0182  & 0.9097 $\pm$ 0.0566 & \textcolor{blue}{0.9861 $\pm$ 0.0052} \\
        \midrule
        Average                 & 0.8321 $\pm$ 0.0159  & 0.849 $\pm$ 0.0110   & 0.9788 $\pm$ 0.0019 \\
        \bottomrule
    \end{tabular}
    \label{tab:t2}
\end{table}

\paragraph{Summary.}
Across all settings, leveraging auxiliary information to define graph neighborhoods significantly improves lesion classification over purely image-based baselines. The full-weighted graph attains the strongest overall performance; a carefully chosen thresholded graph (at $T{=}0.7$) matches it closely while using $\sim25\%$ of the edges.

\section{Discussion} \label{sec:discussion}

This work investigated the integration of population-graph learning with dermoscopic image analysis, where imaging-derived node representations were augmented with auxiliary metadata and structured through weighted edges to model cohort relationships. Representing the dataset as a sparse weighted graph $G=(V,E,W)$ enabled semi-supervised label propagation across the cohort, allowing the learning process to exploit dependencies that are not recoverable from pixel intensities alone. In the context of eight-class lesion classification on ISIC 2019, our results demonstrated that neighborhoods constructed from metadata and geometric scale information consistently improved performance over image-only baselines (\textit{cf.}\ \autoref{tab:results}, \autoref{tab:t3}, \autoref{tab:t2}).  

A central aspect of this study concerned the design of neighborhoods in the adjacency matrix $W$ (\autoref{eq:parisot}). We operationalized metadata similarity through a structured definition of $\gamma(\cdot,\cdot)$, incorporating principled normalization for continuous variables (\autoref{eq:norm}) alongside Kronecker-delta treatment of categorical fields. The empirical evaluation indicated that fully weighted graphs achieved the strongest aggregate performance, while thresholded variants with $T{=}0.7$ preserved most of these gains using only a fraction of edges (\autoref{tab:thresholds}). This finding suggests that sparse neighborhoods, when carefully designed, can retain near-optimal accuracy while reducing graph density, consistent with prior work on medical population graphs \cite{parisot2018disease}.  

Two practical implications emerge for real-world deployment. First, the choice of node features is critical: representations derived from models fine-tuned on dermoscopy data (ISIC initialization) consistently outperformed features transferred solely from generic natural image pretraining (ImageNet). Second, the semantics of edges matter substantially. Graphs without informative neighborhood structure—whether defined identically or at random—performed markedly worse, illustrating that observed improvements stem not merely from model capacity but from the meaningful incorporation of metadata into graph construction.  

This work also has limitations. Chief among them is the reliance on ISIC 2019 as the primary dataset. An ideal benchmark would combine sufficiently large sample size, richly annotated metadata, and standardized protocols for fair comparison across methods. While other population-graph approaches have been validated on large-scale neuroimaging cohorts such as ABIDE \cite{di2014autism} and ADNI \cite{jack2008alzheimer}, the combination of domain expertise, data accessibility, and the need to address dermatology-specific challenges motivated our focus on dermoscopy.  

A second limitation is the absence of a curated, held-out test set with expert-verified lesion masks and ruler annotations, which would provide external validation of the full end-to-end pipeline. Assembling such a dataset remains an open task requiring coordinated expert annotation.

As we did not conduct end-to-end external validation on a held-out clinical cohort with real, in-distribution ruler images, generalization across devices, acquisition protocols, and clinics remains unverified. While experiments on ISIC 2019 demonstrated that our method is effective and yields improved results, the absence of a cohort with expert-verified annotations spanning the entire pipeline—specifically, expert labels for the ruler-derived pixel-to-millimeter scale and geometric features in millimeters used to construct the graph and its edges—limited our ability to explore alternative graph-construction strategies and to further increase accuracy; however, this did not alter the evidence that the graph component positively impacts performance. Assembling such a dataset remains an open task requiring coordinated expert annotation.

Finally, robustness assessments would benefit from extending evaluation to broader graph-learning benchmarks, including PubMed \cite{sen2008collective}, PPI \cite{zitnik2017predicting}, and Reddit \cite{hamilton2017inductive}, which have served as canonical references for evaluating GNN generalization in non-medical domains.  

Several avenues for future research follow naturally from the present findings. With respect to node features, an important direction is to replace static CNN-derived embeddings with representations learned jointly through autoencoder bottlenecks optimized end-to-end with the GNN classifier, as explored in related graph-based medical imaging studies \cite{parisot2018disease}. Alternative backbones such as ResNet, DenseNet, or MobileNet families could also be investigated for their differing trade-offs between accuracy and efficiency.  

Regarding edge weights, our formulation in \autoref{eq:parisot} treated metadata channels uniformly; optimization strategies such as metaheuristic search or differentiable weighting could allow the relative contribution of each metadata field to be learned during training. Robust similarity measures also warrant exploration: although explicit feature-based similarity was avoided to reduce error propagation from noisy embeddings, bounded or regularized kernels may offer a means of capturing finer-grained relationships while mitigating noise sensitivity.  

For numerical metadata, sensitivity to outlier clipping and the normalization function $Norm(\cdot)$ in \autoref{eq:norm} suggests the value of adaptive distribution-aware approaches to further stabilize edge weights. Finally, the GNN architecture itself was fixed to a strong baseline in this study. Systematic evaluation of architectural depth, hidden width, activation functions, normalization schemes, and regularization strategies (\textit{e.g.}, dropout, edge dropout, MixUp for graphs) could provide further gains, while extending comparisons to spatial as well as spectral GNN variants would contextualize the benefits of our design.  

\section{Conclusion} \label{sec:conclusion}

We have presented a population-graph framework for multiclass dermoscopic lesion classification that fuses imaging features with auxiliary metadata and explicit physical scale within a unified GNN. Experiments on ISIC 2019 demonstrated that metadata-informed neighborhoods significantly improve precision, recall, and AUC over image-only baselines, with fully weighted graphs performing best and thresholded graphs at $T{=}0.7$ achieving nearly identical accuracy while requiring far fewer edges (\autoref{tab:results}, \autoref{tab:thresholds}).  

\begin{figure}[!h]
    \centering
    \begin{subfigure}{0.49\textwidth}
        \centering
        \includegraphics[width=\textwidth]{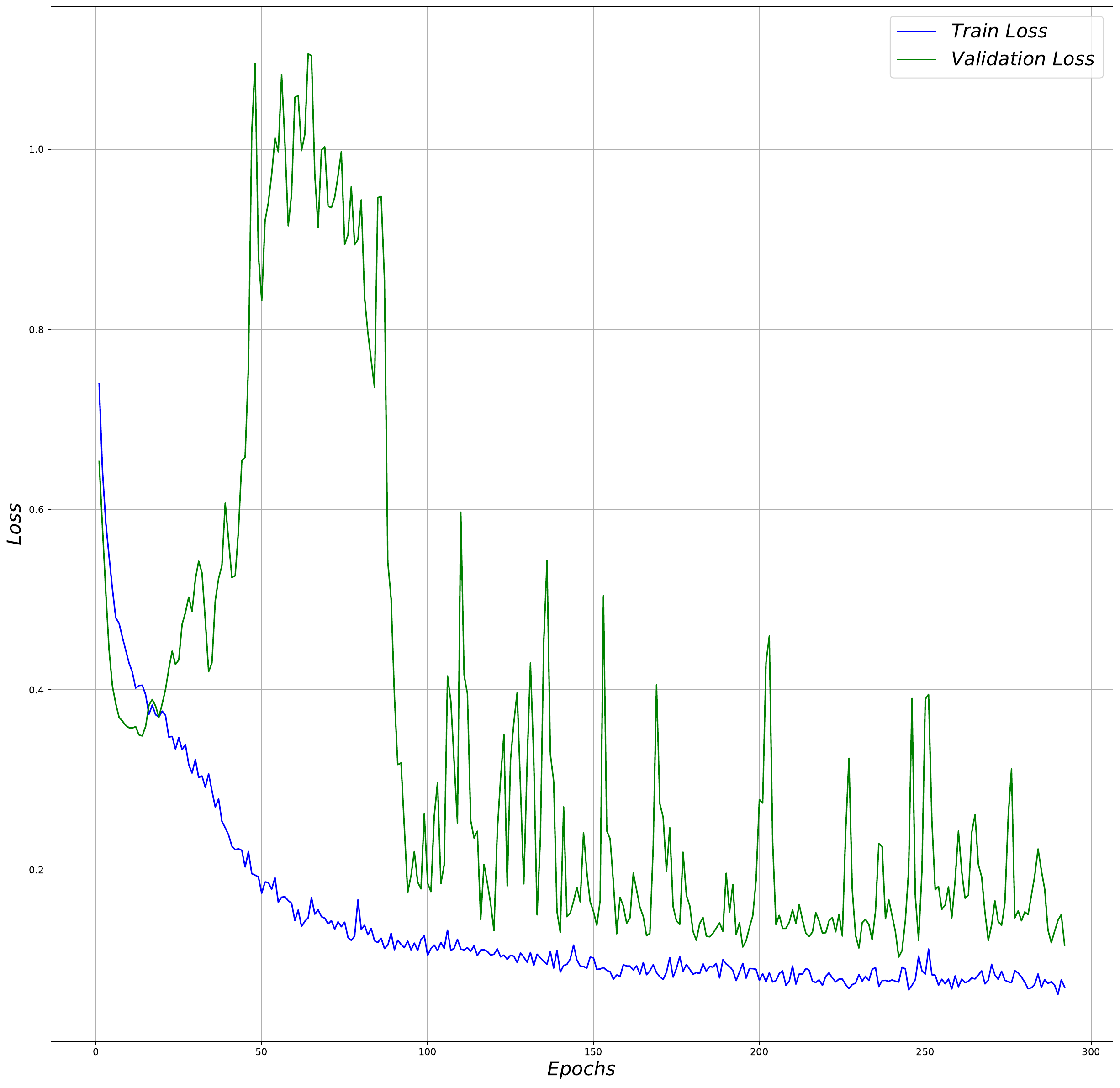}
    \end{subfigure}
    \hfill
    \begin{subfigure}{0.49\textwidth}
        \centering
        \includegraphics[width=\textwidth]{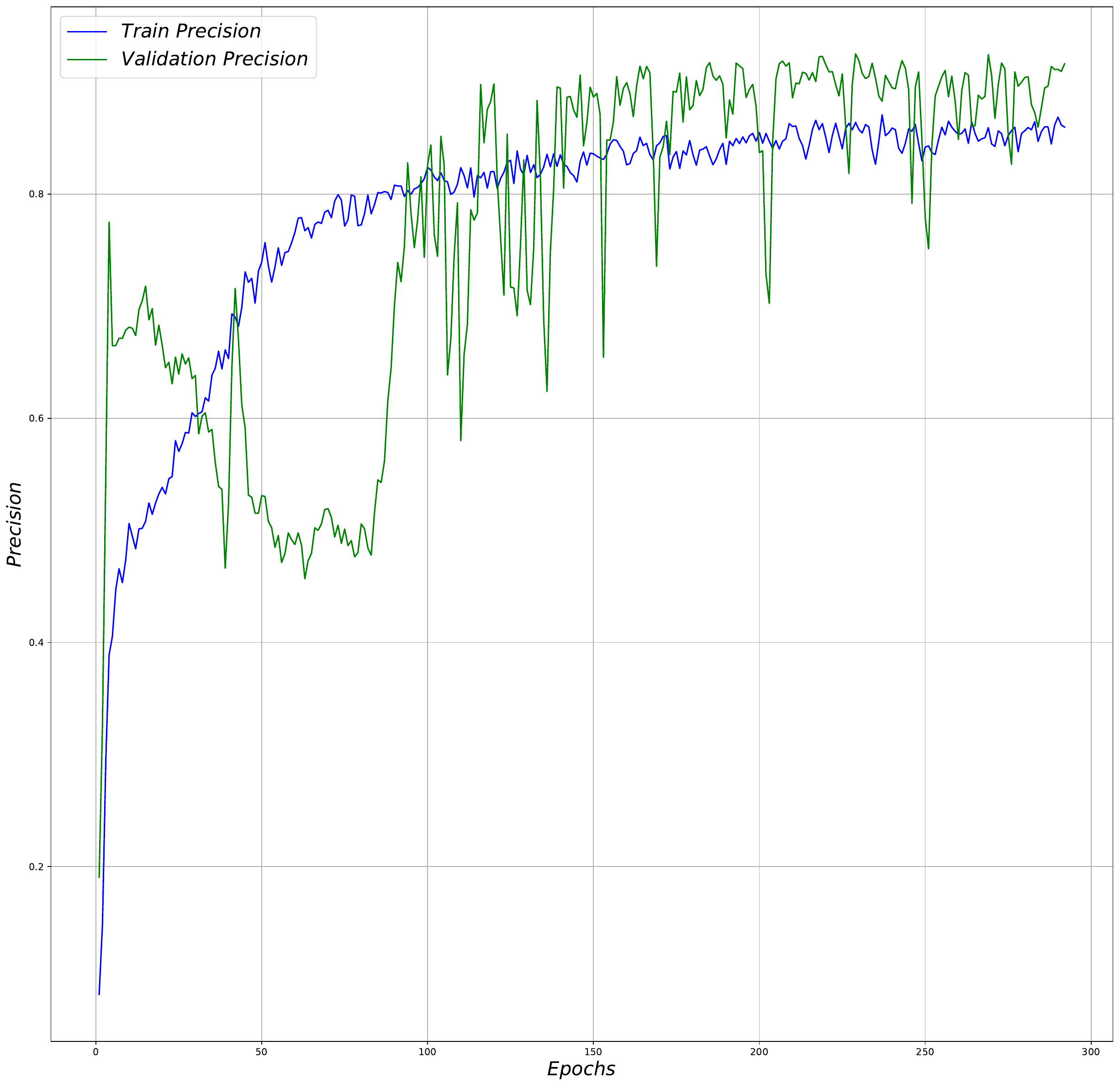}
    \end{subfigure}
    \hfill
    \begin{subfigure}{0.49\textwidth}
        \centering
        \includegraphics[width=\textwidth]{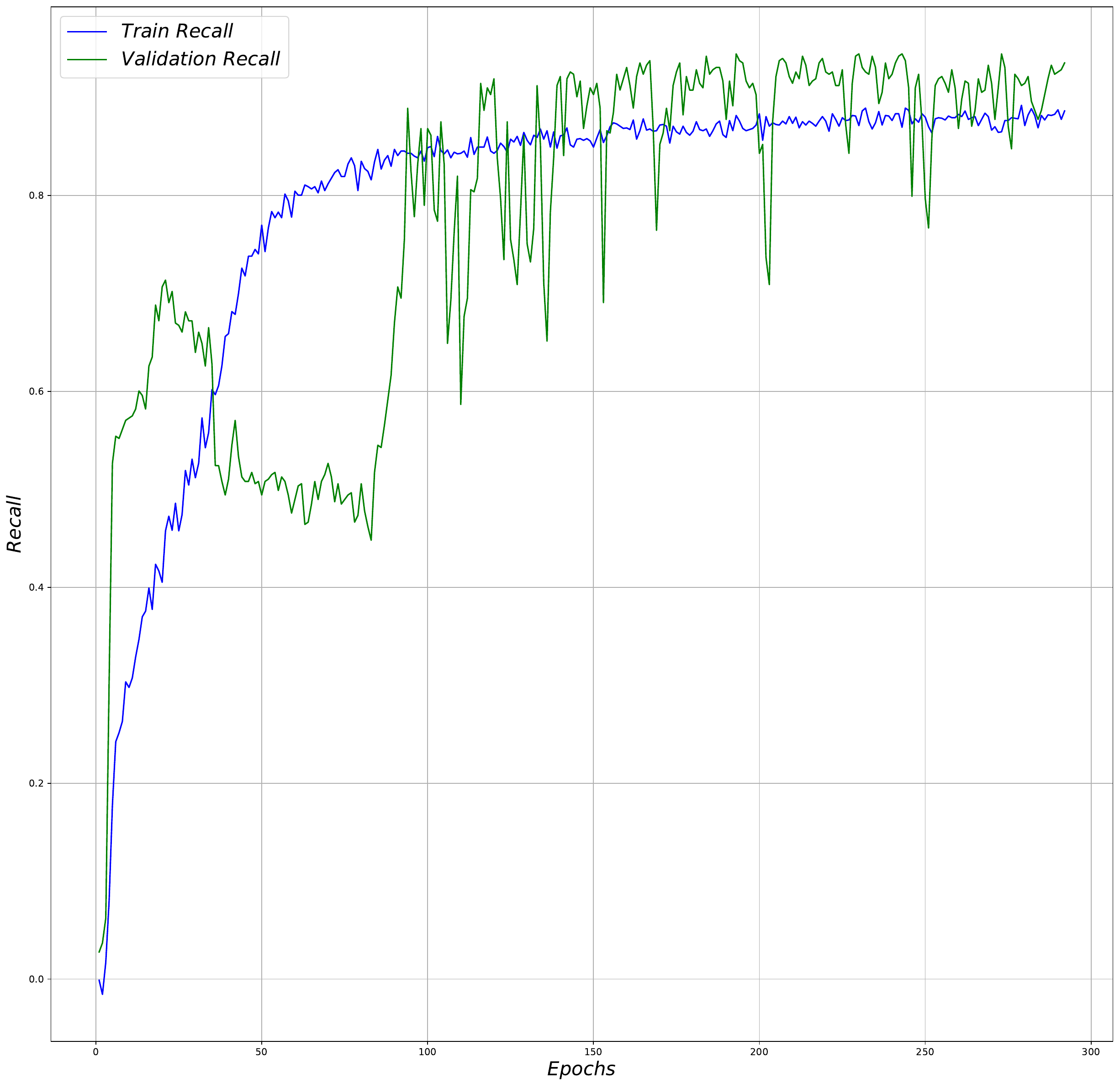}
    \end{subfigure}
    \hfill
    \begin{subfigure}{0.49\textwidth}
        \centering
        \includegraphics[width=\textwidth]{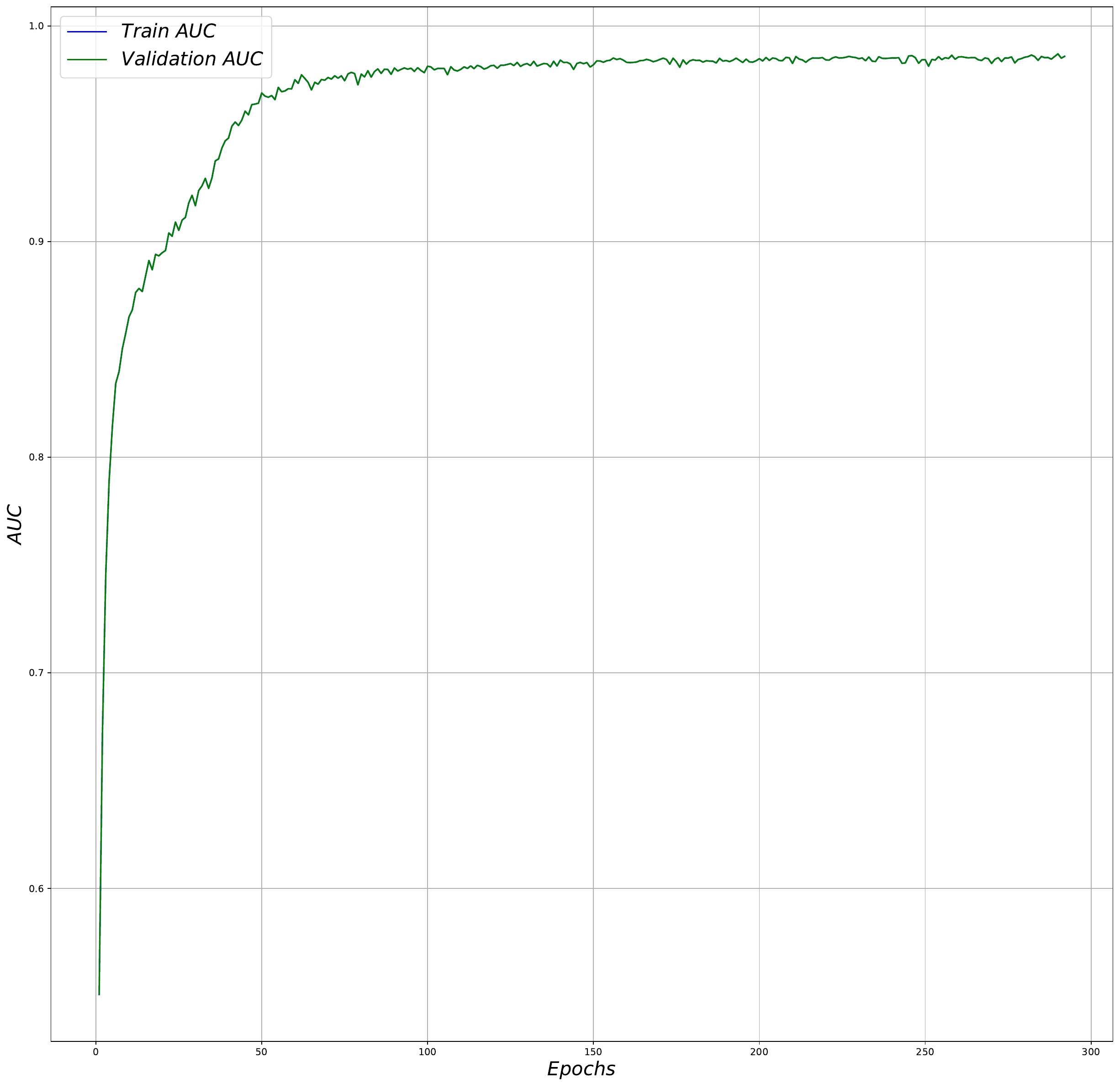}
    \end{subfigure}
    \caption{
        Training/validation curves for the full-weighted edge construction (ISIC 2019–initialized node features): loss, precision, recall, and AUC across epochs.
    }
    \label{fig:epochs-weighted}
\end{figure}

\begin{figure}[!h]
    \centering
    \begin{subfigure}{0.49\textwidth}
        \centering
        \includegraphics[width=\textwidth]{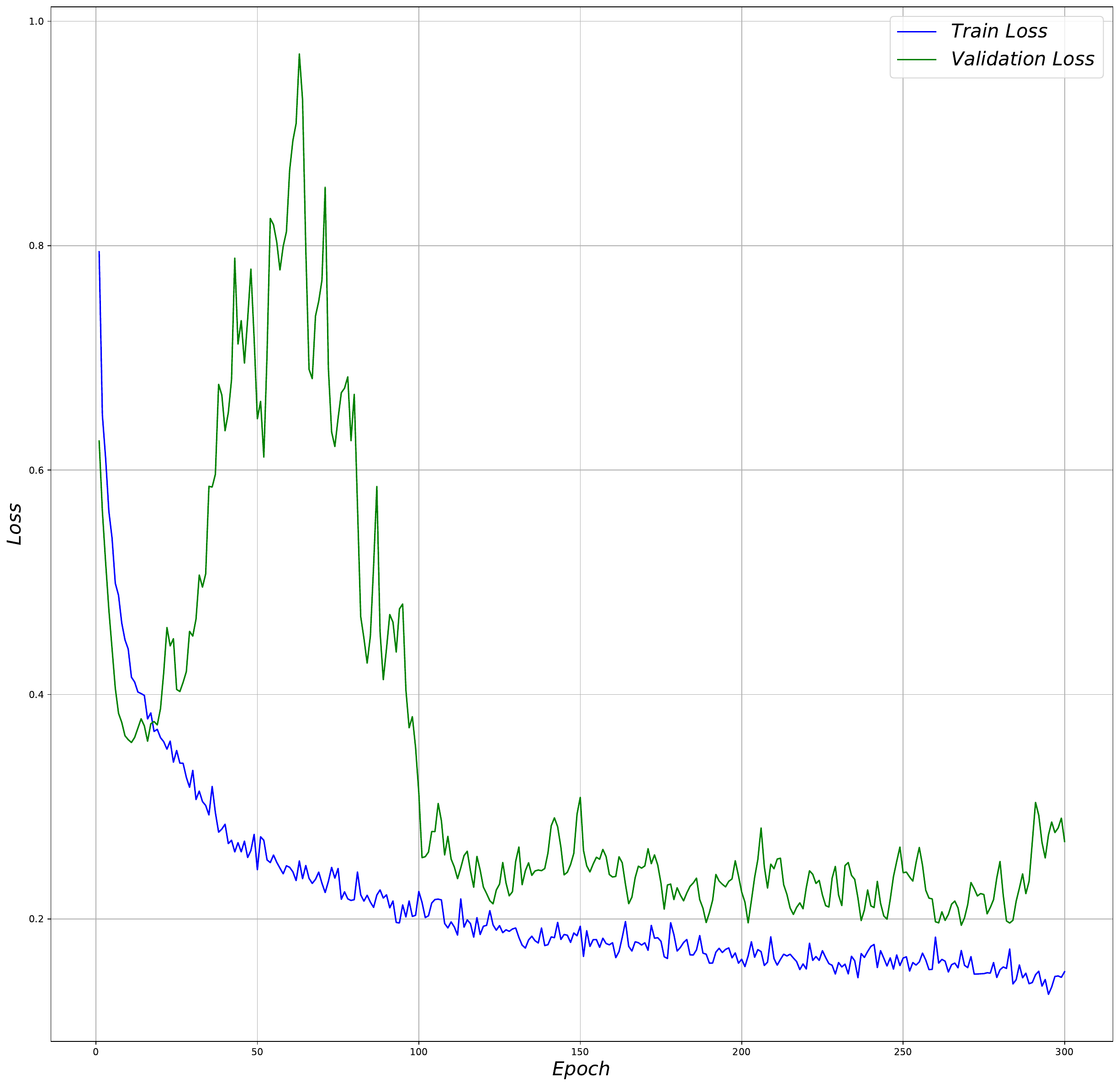}
    \end{subfigure}
    \hfill
    \begin{subfigure}{0.49\textwidth}
        \centering
        \includegraphics[width=\textwidth]{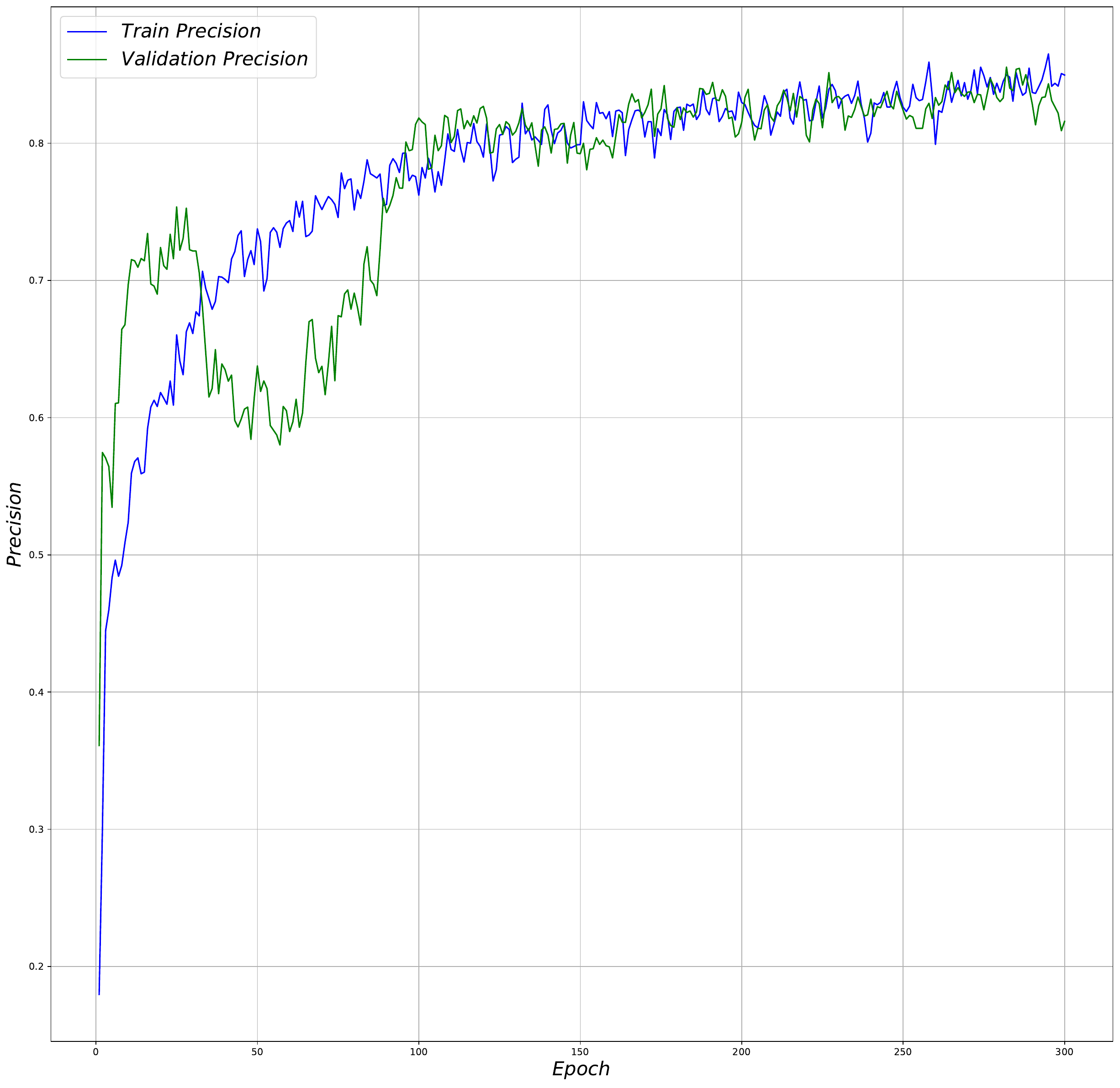}
    \end{subfigure}
    \hfill
    \begin{subfigure}{0.49\textwidth}
        \centering
        \includegraphics[width=\textwidth]{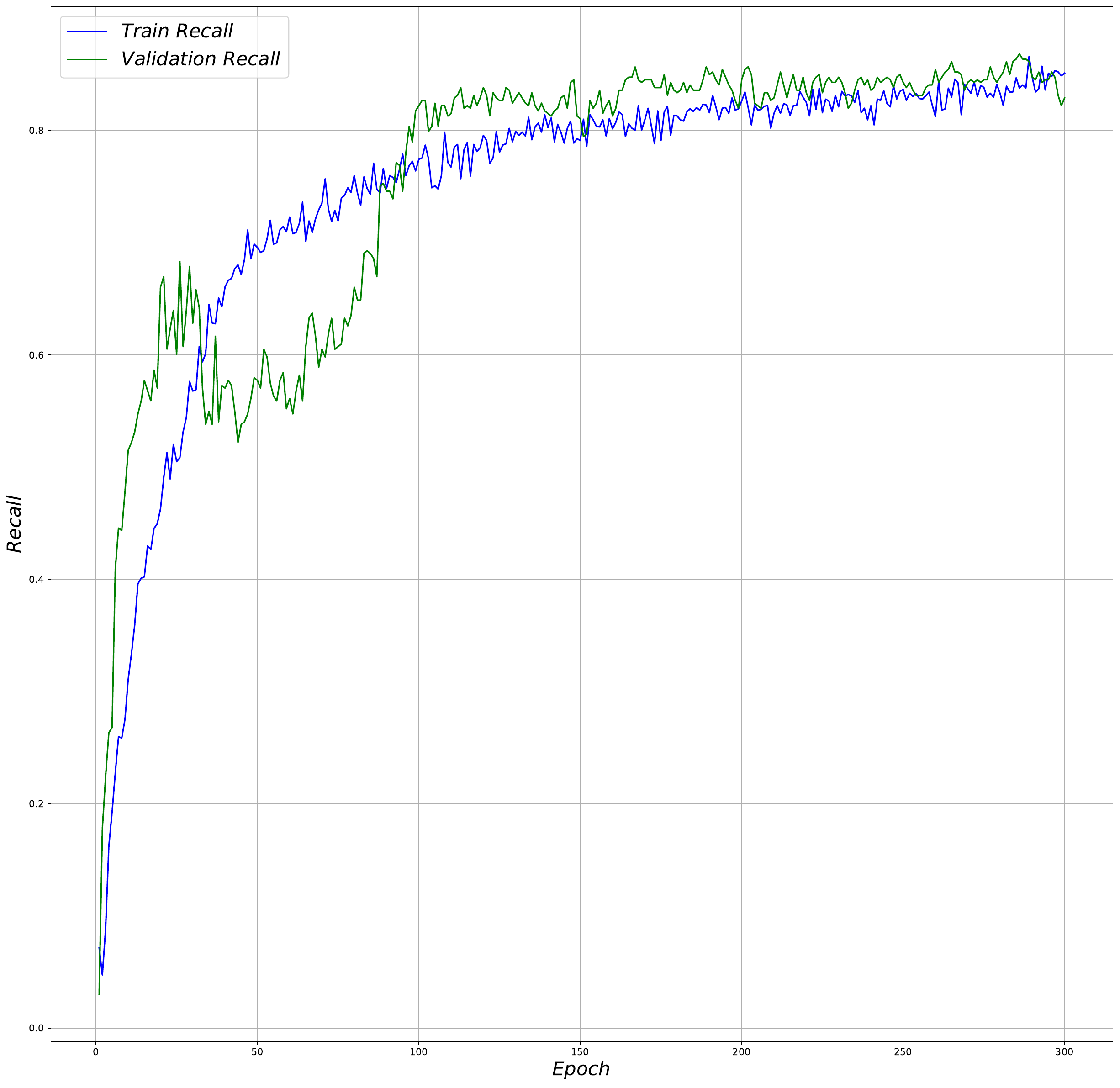}
    \end{subfigure}
    \hfill
    \begin{subfigure}{0.49\textwidth}
        \centering
        \includegraphics[width=\textwidth]{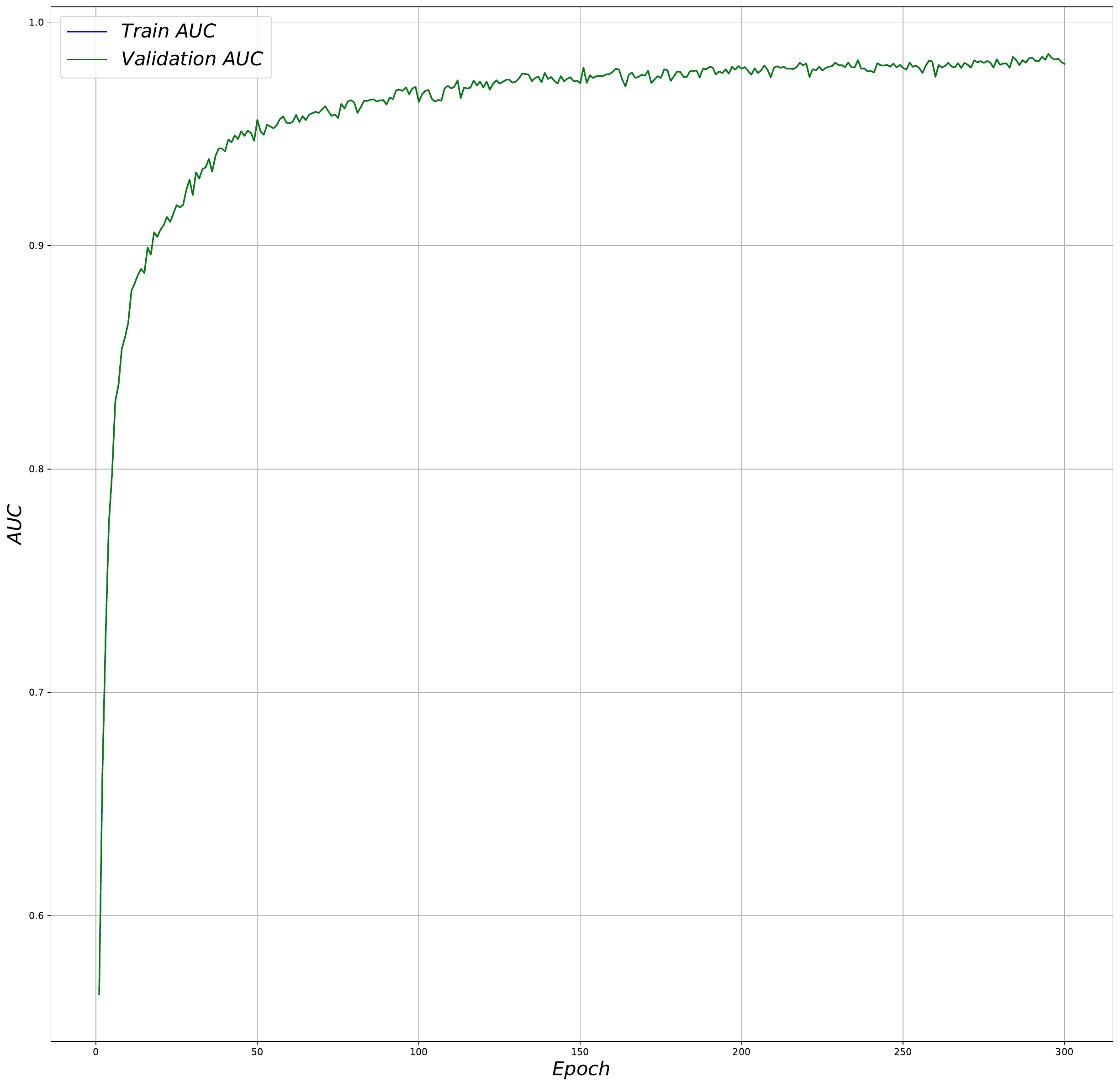}
    \end{subfigure}
    \caption{
        Training/validation curves for the thresholded edge construction (ISIC 2019–initialized node features): loss, precision, recall, and AUC across epochs.
    }
    \label{fig:epochs}
\end{figure}

These improvements, together with stable training behavior (\suppfig{fig:epochs-weighted}, \suppfig{fig:epochs}) and high-quality segmentation and pixel-scale estimation, underscore the promise of graph-based learning for dermoscopic analysis. Looking forward, the development of curated test sets, particularly with real, expert-verified lesion masks and ruler annotations, along with external validation on clinical cohorts, remains essential steps toward translating graph neural networks into clinical dermatology. The integration of learned edge semantics, and systematic benchmarking against broader datasets, will further establish the robustness and transportability of this framework.

\label{sec:acknowledgement}
\section*{Acknowledgement}
The authors would like to thank the doctors and medical staff at Razi Hospital, Tehran University of Medical Sciences, Tehran, Iran, as well as Dr. Farideh Beyki. The authors also acknowledge the use of \emph{ChatGPT} (\emph{OpenAI}, \href{https://chatgpt.com/}{chatgpt.com}) for assistance with language editing and improving the readability of the manuscript. The research ideas, analysis, and conclusions are entirely the responsibility of the authors.

\bibliographystyle{unsrtnat}
\bibliography{references}  

\beginsupplementary
\newpage
\label{sec:appendic}
\section{Supplementary Material}

\begin{table}[!h]
    \centering
    \caption{
        Class distribution in the ISIC 2019 dataset, along with the number of samples allocated to the training and validation sets used for training the feature-extraction model.
    }
    \small
    \begin{tabular}{lrrr}
        \toprule
        Classes                 & Total & Train & Validation \\
        \midrule
        Melanoma                & 4522  & 2586  & 601  \\
        Melanocytic Nevus       & 12875 & 8514  & 2168 \\
        Basal Cell Carcinoma    & 3323  & 2027  & 511  \\
        Actinic Keratosis       & 867   & 480   & 113  \\
        Benign Keratosis        & 2624  & 1446  & 353  \\
        Dermatofibroma          & 239   & 126   & 39   \\
        Vascular Lesion         & 253   & 168   & 44   \\
        Squamous Cell Carcinoma & 628   & 354   & 97   \\
        \bottomrule
    \end{tabular}
    \label{table:isic2019-trainval}
\end{table}

\begin{table}[!h]
    \centering
    \caption{
        Details and parameters of the graph neural network model architecture and training.
    }
    \small
    \begin{tabular}{|ll|ll|}
        \hline
        Model           & GCN \cite{kipf2016semi}   & Loss              & Weighted Categorical Cross-Entropy \\
        Batch Size      & 256                       & Validation        & 5-Fold Cross-Validation \\
        \hline
        Optimizer       & Adam                      & \# Epochs         & 300 \\
        Learning-Rate   & 0.01                      & \# Hidden Layers  & 32 \\
        \hline
    \end{tabular}
    \label{table:parameters}
\end{table}

\begin{table}[!h]
    \centering
    \caption{
        Sample count and loss function weights for all classes.
    }
    \small
    \begin{tabular}{lllllllll}
        \toprule
        i          & 0      & 1      & 2      & 3      & 4      & 5      & 6      & 7      \\
        \midrule
        \# Samples & 400    & 400    & 400    & 274    & 400    & 74     & 41     & 177    \\
        $w^{+}_i$    & 2.7075 & 2.7075 & 2.7075 & 3.9525 & 2.7075 & 14.635 & 26.414 & 6.1186 \\
        $w^{-}_i$   & 0.6132 & 0.6132 & 0.6132 & 0.5724 & 0.6132 & 0.5176 & 0.5096 & 0.5444 \\
        \bottomrule
    \end{tabular}
    \label{table:weights}
\end{table}

\end{document}